\pdfoutput=1

\documentclass[runningheads]{llncs}

\usepackage{eccv}

\usepackage{eccvabbrv}
\usepackage{graphicx}
\usepackage{booktabs}
\usepackage[accsupp]{axessibility}
\usepackage{array}
\usepackage{placeins}

\graphicspath{{figures/}}

\usepackage[breaklinks,colorlinks,allcolors=eccvblue,bookmarksnumbered]{hyperref}
\usepackage{xurl}   % break long URLs in the bibliography

\usepackage{orcidlink}
\usepackage{fontawesome5}

\newcommand{\syngallery}{SynGallery}
\newcommand{\GAP}{\mathrm{GAP}}
\newcommand{\GAPm}{\mathrm{GAP}^{-}} % GAP without distractors

\newcolumntype{M}{>{\centering\arraybackslash}m{0.135\linewidth}}
\newcolumntype{N}{>{\centering\arraybackslash}m{0.15\linewidth}}

\hypersetup{
  pdftitle={SynGallery: A Synthetic Gallery of Real Paintings for Instance-Level Artwork Recognition},
  pdfauthor={Patryk Bartkowiak, Jakub Markil, Bartosz Kotrys, Dominik L. Michels, Soeren Pirk, Wojtek Palubicki},
  pdfkeywords={computer vision, synthetic data, art painting retrieval}
}

\begin{document}

% ---------------------------------------------------------------
% Paper metadata
\title{SynGallery: A Synthetic Gallery of Real Paintings for Instance-Level Artwork Recognition}
\titlerunning{SynGallery}

\author{
    Patryk Bartkowiak\inst{1,5}\orcidlink{0009-0005-6659-8302} \and
    Jakub Markil\inst{1}\orcidlink{0009-0008-4914-4458} \and
    Bartosz Kotrys\inst{2} \and
    Dominik L. Michels\inst{3,5}\orcidlink{0000-0002-1621-325X} \and
    S\"oren Pirk\inst{4,5}\orcidlink{0000-0003-1937-9797} \and
    Wojtek Pa{\l}ubicki\inst{1,5}\orcidlink{0000-0002-2374-346X}
}
\authorrunning{Bartkowiak et al.}

\institute{
    Adam Mickiewicz University \and
    ArtiCollect \and
    KAUST \and
    Kiel University \and
    GreenMatterAI
}

\maketitle

% Public code and dataset (arXiv version only).
    \begin{center}
    \small
    \textbf{\href{https://github.com/SynGallery/syngallery-bench}{\faGithub~Code}}
    \quad \textbar \quad
    \textbf{\href{https://huggingface.co/collections/patryk-bartkowiak/syngallery}{\faDatabase~Dataset}}
    \end{center}

% ---------------------------------------------------------------
% Paper sections
\begin{abstract}
Instance-level artwork recognition requires matching a handheld visitor photograph to a specific work in a large museum collection. This is challenging because painting datasets typically provide clean catalog images for training, while test queries are captured under oblique viewpoints, gallery lighting, reflections, frames, and other scene-level variations. We present SynGallery, a synthetic gallery dataset for artwork retrieval that addresses this gap without collecting additional real photographs. Starting from catalog images of real paintings, we place each artwork into a procedurally generated 3D gallery scene and render it from multiple viewpoints under varied geometric and appearance conditions, while preserving the exact identity of the original work. The resulting dataset contains 24,490 rendered views of 4,898 paintings from the Met benchmark. We show that these synthetic views provide a stronger training signal than the corresponding studio photographs. At the same number of training data points, training only on SynGallery improves art painting recognition from 67.18 to 73.47 GAP$^-$. When added to the full Met training set, SynGallery improves the published benchmark protocol from 35.97 to 38.48 GAP. Ablation experiments show that the gain comes from scene-level view variation rather than photographic realism: reducing the five rendered viewpoints to a single frontal view removes most of the improvement, while simulating capture artifacts such as blur, sensor noise, and image compression consistently reduces performance.
\end{abstract}

\section{Introduction}
\label{sec:intro}

\begin{figure}[t]
  \centering
  \includegraphics[width=\linewidth]{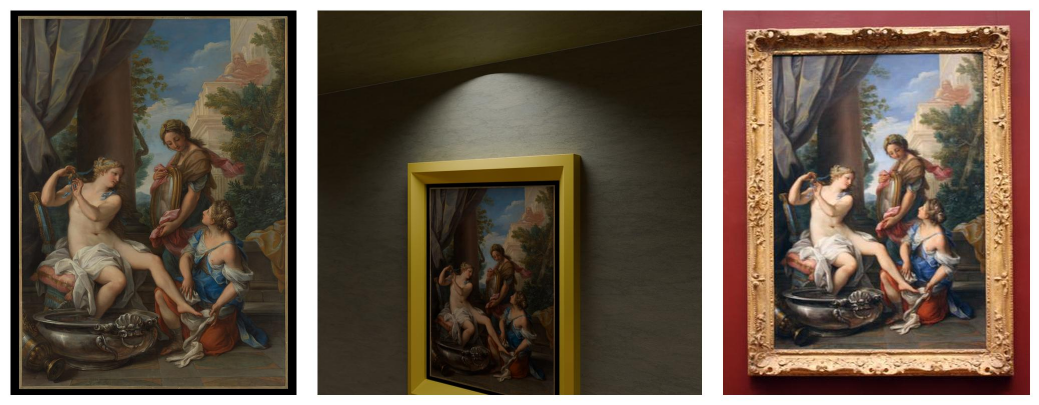}
  \caption{One painting, three types of photos. The Met benchmark trains on studio catalog photos (left) and is tested with handheld visitor photos (right). SynGallery (center) places the scanned catalog into a synthetic 3D gallery, giving the model training views of each painting on a gallery wall---from angles and lighting that studio photos never show.}
  \label{fig:teaser}
\end{figure}

Museum visitors increasingly use mobile devices to identify artworks in real time. This presents a challenging instance-level recognition problem: matching a single query photo to a specific artwork among hundreds of thousands of candidates. The Met dataset~\cite{met_dataset} benchmarks this exact setting. It provides training data of studio catalog photographs of museum exhibits, but evaluates on real photos taken by visitors in the galleries, mixed with thousands of distractor images. This task is difficult for two reasons: (1) most artworks have very few catalog photos (over 60\% have just one), and (2) the studio training images look nothing like the test ones: clean, frontal studio scans versus handheld shots taken at angles, under gallery lighting, and often through glass.

This difference between studio and visitor photos is the core challenge when training vision models, and collecting many real visitor photos per artwork is impractical at scale. However, since museums already have digitized catalogs of their collections, we procedurally generate a 3D gallery scene for each painting (see Fig.~\ref{fig:teaser})---placed on a wall inside a frame, sometimes behind glass, next to a placard, on varied floors and lighting, rendered from different view points. These renders can be used to train vision models with the scene-level variation that studio scans lack, without collecting a single real-world visitor photo. We introduce this dataset as \syngallery{}.

The renders are not necessarily meant to have a high perceptual similarity to real photos. Instead, we want to improve performance of vision models on downstream tasks. In fact, embedding analysis using DINOv3~\cite{simeoni2025dinov3} reveals that the \syngallery{} renders form their own cluster, no closer to the real visitor photos than the studio scans are. This indicates that it is not realism but variation that enhances the training datasets. By changing the scene attributes, the \syngallery{} training data allows the vision model to learn ignoring the differences between catalog and visitor photos. Throughout the paper we use \emph{view variation} in this scene-level sense: the joint change of viewpoint, lighting, frame, and gallery context under which the same painting is observed. We demonstrate that this variation, with viewpoint coverage as the strongest single factor, drives retrieval performance, whereas forcing the renders to mimic phone camera artifacts actively degrades accuracy.

The \syngallery{} renders serve as effective training data. At an equivalent image budget, training on the synthetic renders alone beats training on the real studio photos. Added to the full real training set, they outperform the model released with the Met benchmark, under its exact protocol---with no change to the method and no extra real data.

We make three primary contributions:
\begin{enumerate}
  \item \syngallery{}, a dataset of 24{,}490 gallery renders of 4{,}898 Met paintings seen from five viewpoints, each saved with its scene details (Sec.~\ref{sec:dataset_method}).
  \item Evidence that the renders are a stronger training signal than the studio photographs they are derived from: at a matched image budget, training on renders alone beats training on the real studio photos (closed painting recognition $\GAPm$ 73.47 vs.\ 67.18), and adding the renders to the full 397k-image training set improves over the model released with the Met benchmark under its exact protocol ($\GAP$ 38.48 vs.\ 36.1) (Sec.~\ref{sec:results}).
  \item We show through a number of ablation studies that the performance improvement comes from scene-level view variation: covering more viewpoints is the key ingredient, the renders form their own domain rather than imitating visitor photos, and simulating low-level capture artifacts (e.g., blur, sensor noise) actively harms performance (Sec.~\ref{sec:results}).
\end{enumerate}

\section{Related Work}
\label{sec:related}

\paragraph{Instance-level recognition and the Met benchmark.}
Instance-level recognition (ILR) aims to match a query to a specific target within a massive database. Traditionally, ILR is benchmarked on architectural and natural landmarks, using datasets like Google Landmarks v2~\cite{weyand2020gldv2}. These benchmarks expose the core difficulties of the task: a highly long-tailed class distribution~\cite{kang2019decoupling} and a high volume of out-of-domain distractor queries. Standard retrieval architectures address this via contrastive learning, typically employing generalized-mean (GeM) pooling with whitening and hard-negative mining~\cite{radenovic2018gem}. The Met dataset brings these specific ILR challenges into the cultural heritage domain. It requires models to match unconstrained visitor photos to clean studio catalog images — a large domain gap made even harder because most artworks have only one training image.

\paragraph{Artwork and museum recognition.}
Recognizing artworks under real-world gallery conditions is usually treated as a domain adaptation problem. Open MIC~\cite{koniusz2018openmic} addressed the studio-to-visitor shift by collecting real photos that capture viewpoint changes, gallery lighting, and glass glare. Other large-scale art datasets, such as OmniArt~\cite{strezoski2017omniart}, focus on multi-task learning for attributes like style or artist attribution rather than specific instance retrieval. Furthermore, while strong foundation models like Vision Transformers (ViTs) have been evaluated for art tasks such as authentication~\cite{schaerf2023art}, these approaches typically evaluate 2D style rather than 3D geometry. Instead of collecting real visitor photos, we use 3D renders to generate new viewpoints and lighting.

\paragraph{Whole-image retrieval vs. detect-and-rectify.}
An alternative to matching raw visitor photos is to first detect the artwork in the query, rectify it to a fronto-parallel view, and retrieve using only the rectified crop. We deliberately follow the end-to-end protocol of the Met benchmark, which operates on unmodified query photographs, for three reasons. First, a detection and rectification stage has its own failure modes (strongly oblique views, occlusions, non-rectangular or unframed works) whose errors propagate to retrieval. Second, several nuisance factors that separate visitor photos from catalog scans, including glass reflections, glare, and gallery illumination, affect the painting surface itself and survive rectification. Third, a single robust embedding handles every query in one forward pass and remains directly comparable to prior results. The two directions are complementary: the same procedural scenes we use for training an embedding could supply supervision for such a detection and rectification stage.

\paragraph{Synthetic data for retrieval and art.}
Simulation-based synthetic imagery has been widely used for training visual models,
ranging from virtual urban environments and domain-randomized object detection to
scalable procedural scene generation~\cite{ros2016synthia,tremblay2018training,
greff2022kubric,raistrick2023infinigen,Cieslak2024gda,kaluzny2024laesileafareaestimation}. Synthetic training data has also been leveraged in cultural heritage contexts, but historically for different tasks. The closest artifact to our pipeline is VirtualGallery~\cite{weinzaepfel2019virtualgallery}, which renders a 3D museum environment strictly to train visual camera localization, not to recognize the artworks themselves. When synthetic data is applied directly to art classification, it is almost exclusively 2D-generative: for example, using diffusion models to generate synthetic forgeries to augment authentication datasets~\cite{ostmeyer2024synthetic}. Closer to our task, Wu~\etal~\cite{wu2026ilr} show that fine-tuning on generated views of synthetic object instances improves instance-level retrieval across several domains; in their setting the instances themselves are newly generated, whereas artwork recognition requires preserving the identity of existing works.

However, 2D generators change the actual content and identity of the artwork. Recent work evaluating the synthetic-to-real domain gap demonstrates that the utility of synthetic training data depends on the non-linear interplay of structural (geometric) consistency and appearance similarity~\cite{bartkowiak2026sadge}. Because instance-level recognition demands exact identity matching, structural integrity must be preserved. Instead, we use 3D domain randomization~\cite{tobin2017dr}: we place the original 2D catalog image into a generated 3D scene. This provides the critical geometric variation (multiple camera poses) and appearance changes (lighting, frames, glass) needed to close the domain gap, while strictly preserving the authentic identity of the cataloged artwork.

\section{Dataset Generation Method}
\label{sec:dataset_method}

\syngallery{} maps each Met catalog image directly onto a virtual canvas and renders it inside a randomized three-dimensional gallery using Blender~5.1. The catalog image is used as the canvas texture, so the artwork's instance-level identity is preserved exactly, while the room, lighting, frame, and viewpoint around it are sampled anew for every scene. This follows the idea of 3D domain randomization, but with an important constraint: because the task is instance-level retrieval, the depicted artwork itself must not change. We therefore keep the catalog image fixed and randomize only the gallery in which it is observed. 
%The dataset comprises 24{,}490 RGB renders --- the 4{,}898 paintings of the Met benchmark observed from five fixed viewpoints each --- generated in Blender~5.1.

\begin{figure}[tbp]
  \centering
  \includegraphics[width=\linewidth]{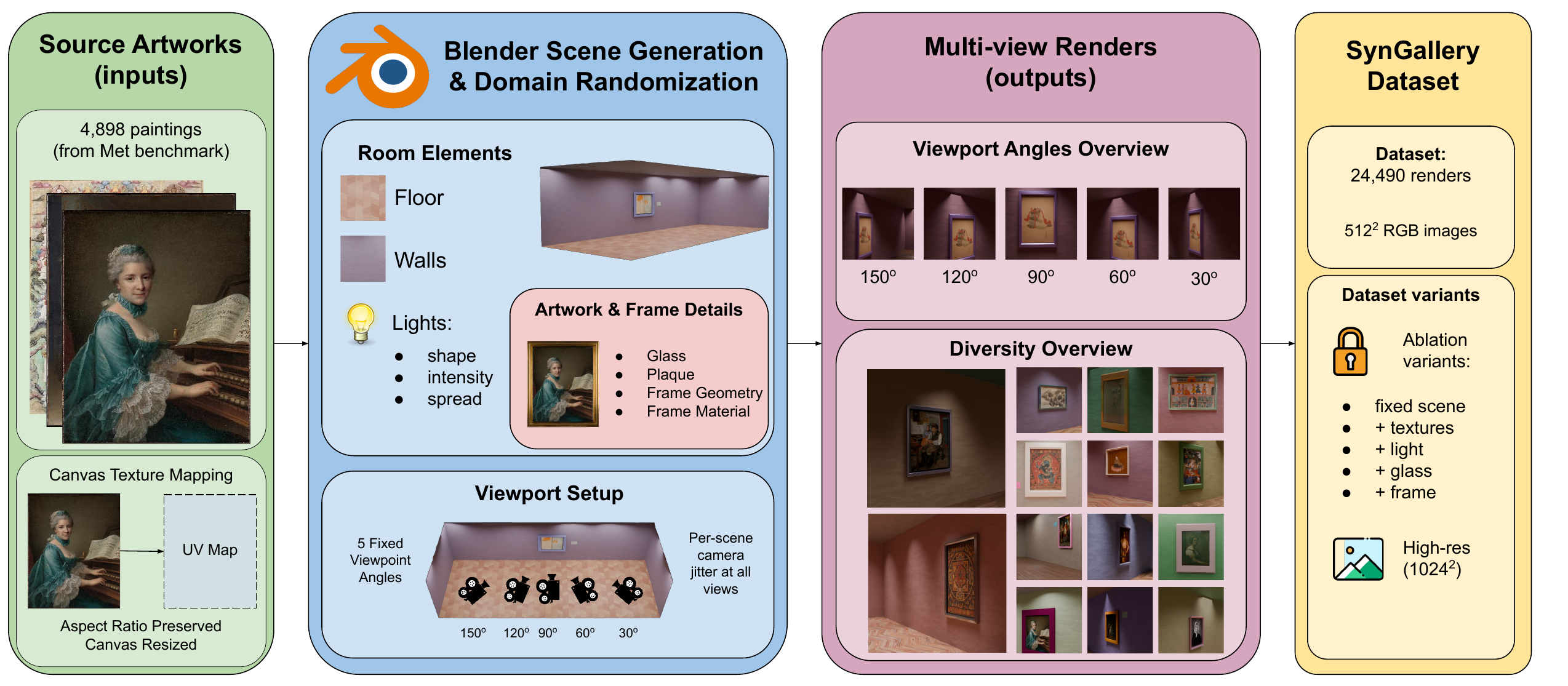}
  \caption{An overview of our procedural model for gallery artwork paintings used to generate the training dataset \syngallery{}. }
  \label{fig:overview}
\end{figure}

\subsection{Identity-Preserving Gallery Rendering}

\begin{figure}[tbp]
    \centering
    \setlength{\tabcolsep}{2pt}
    \begin{tabular}{MMMMMM}
        Source image & $150^\circ$ & $120^\circ$ & $90^\circ$ & $60^\circ$ & $30^\circ$ \\
        \includegraphics[width=\linewidth]{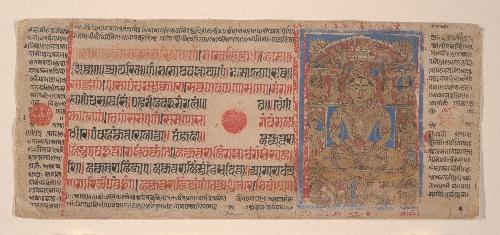} &
        \includegraphics[width=\linewidth]{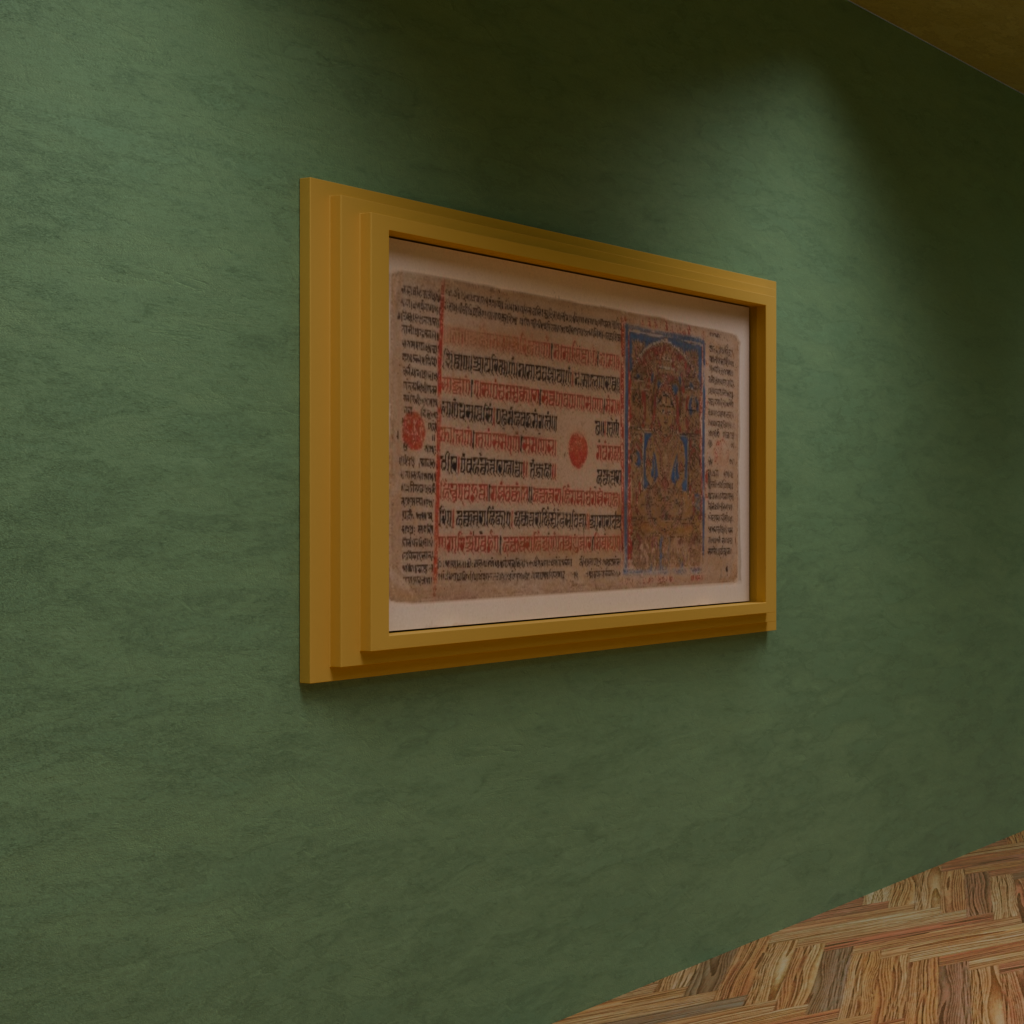} &
        \includegraphics[width=\linewidth]{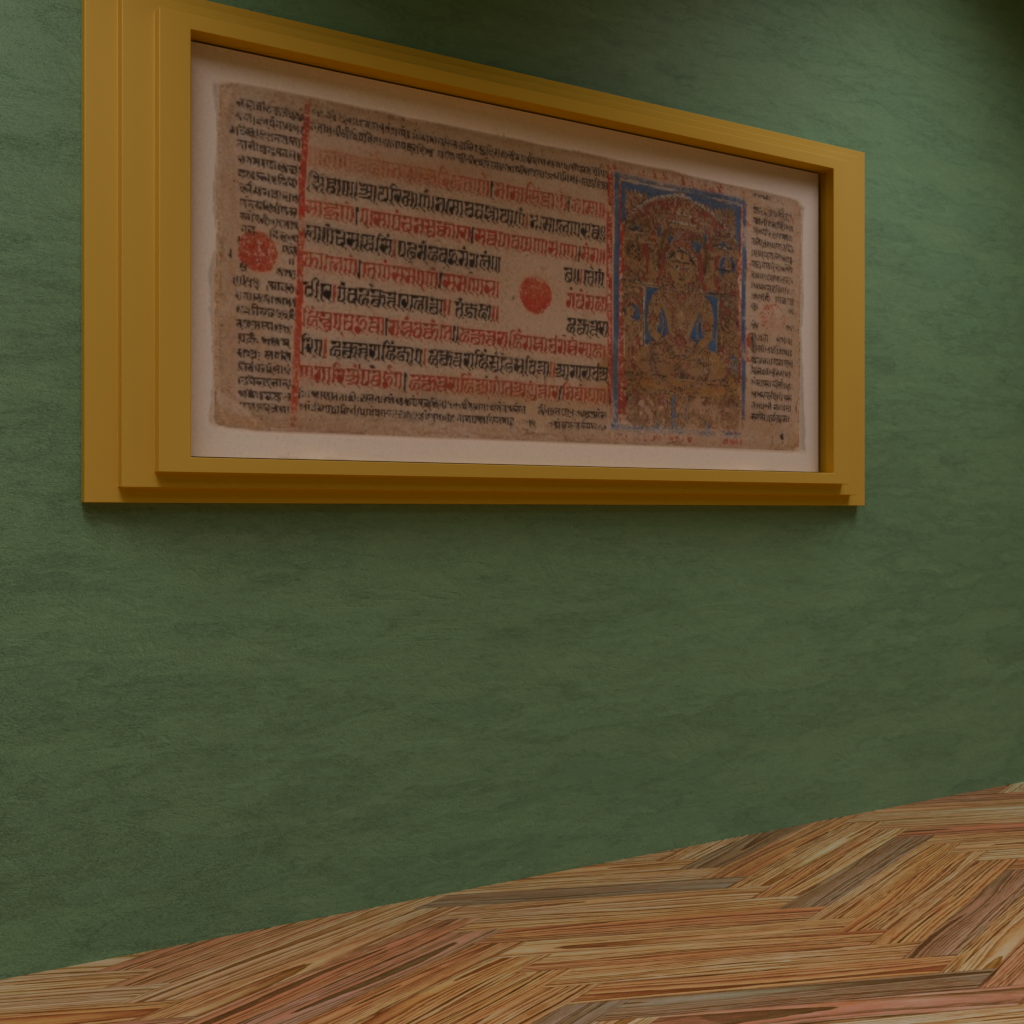} &
        \includegraphics[width=\linewidth]{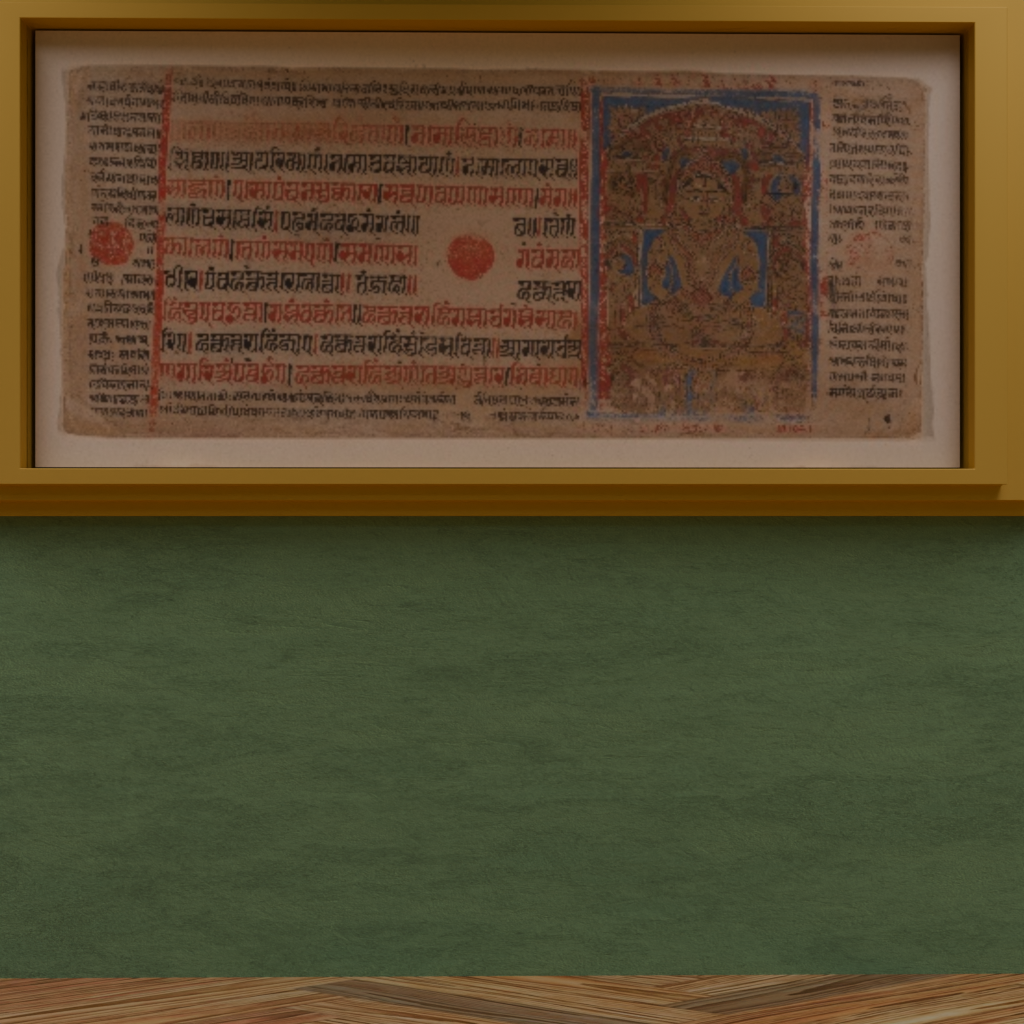} &
        \includegraphics[width=\linewidth]{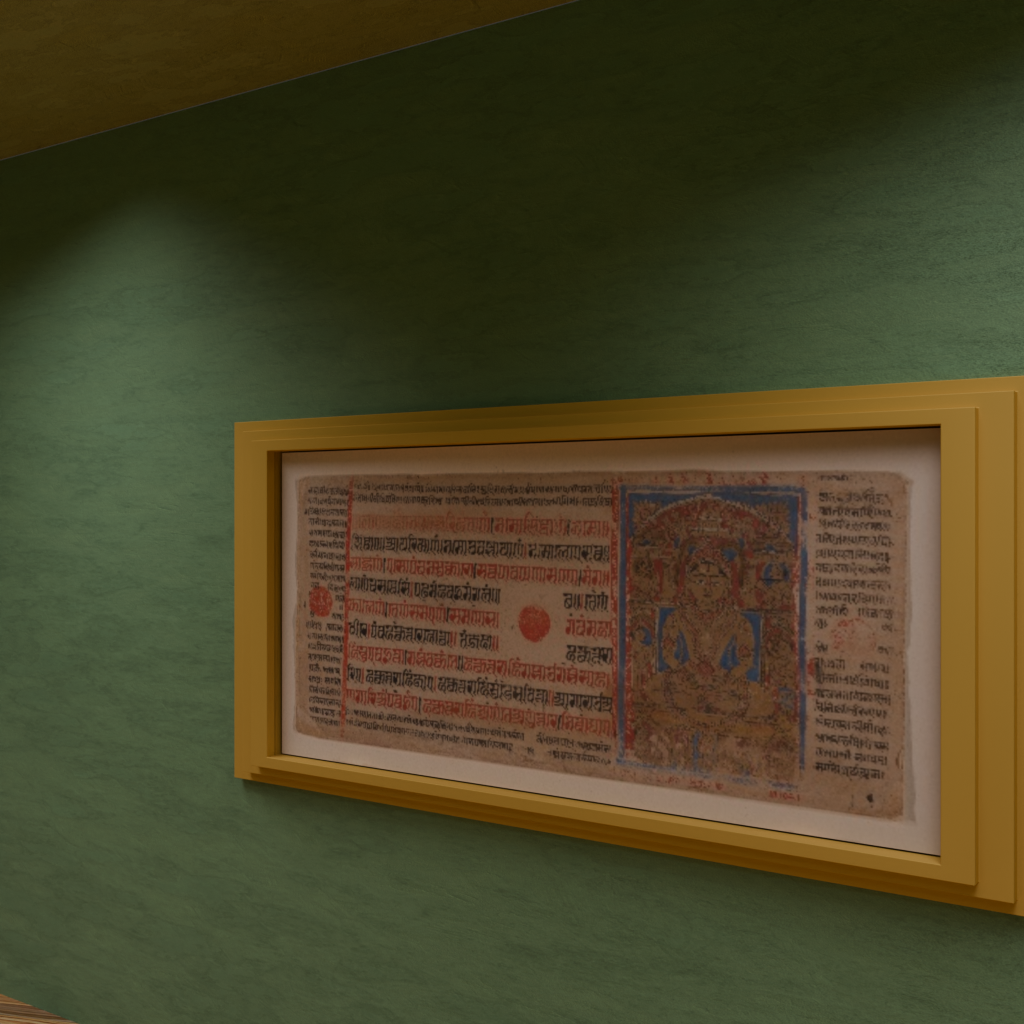} &
        \includegraphics[width=\linewidth]{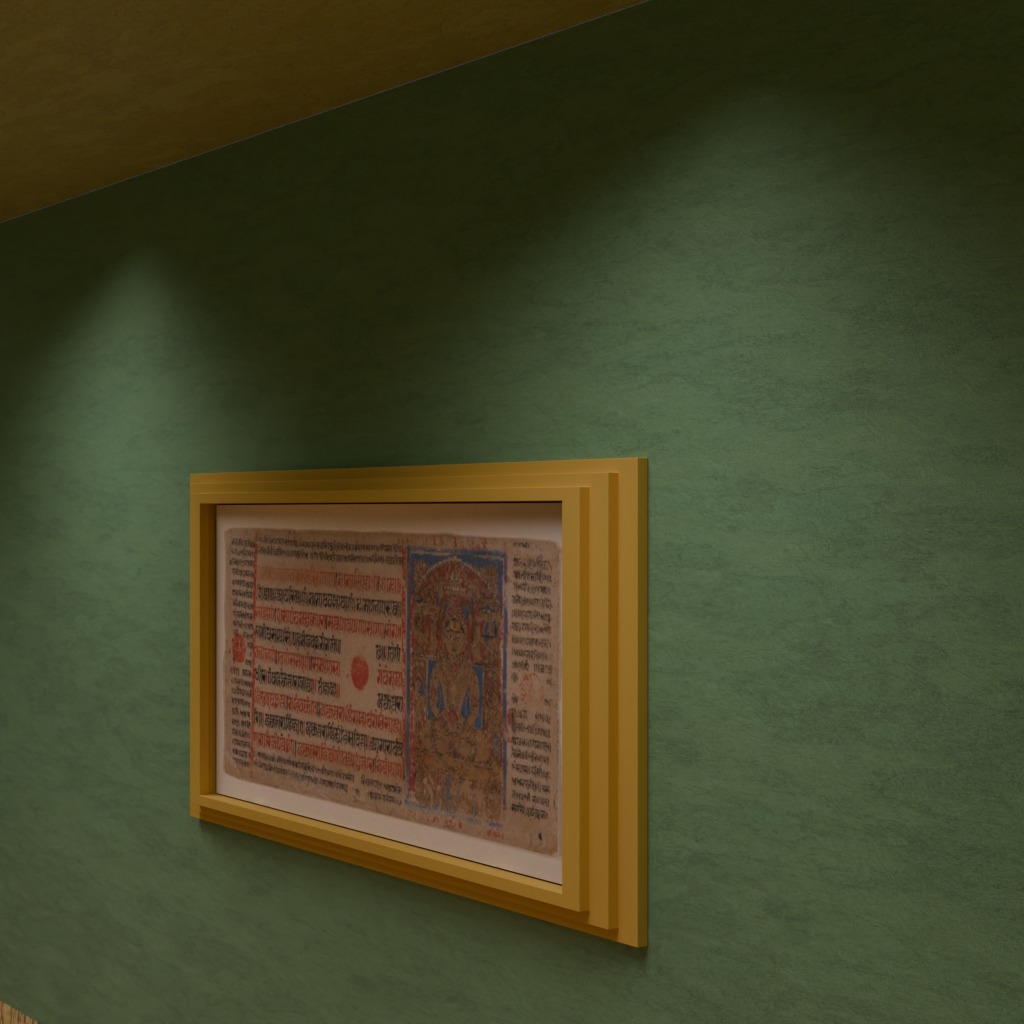} \\
        \includegraphics[width=\linewidth]{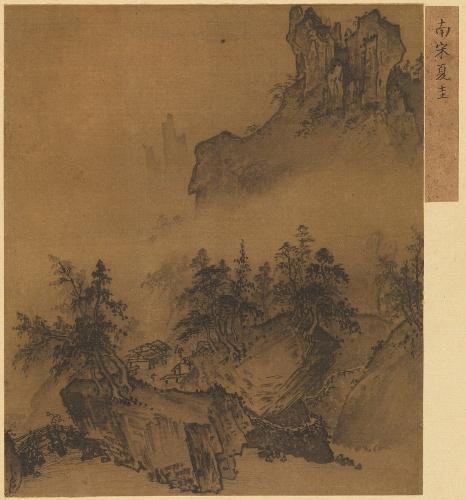} &
        \includegraphics[width=\linewidth]{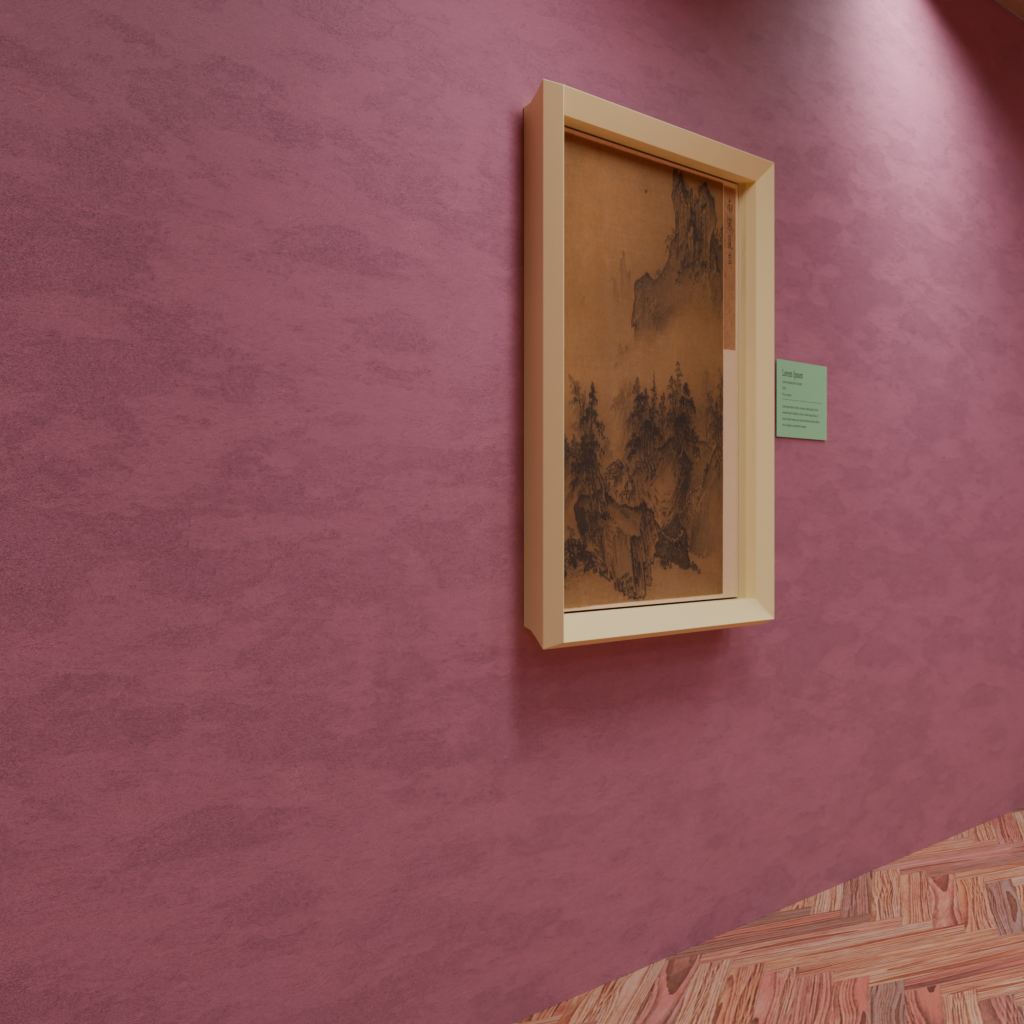} &
        \includegraphics[width=\linewidth]{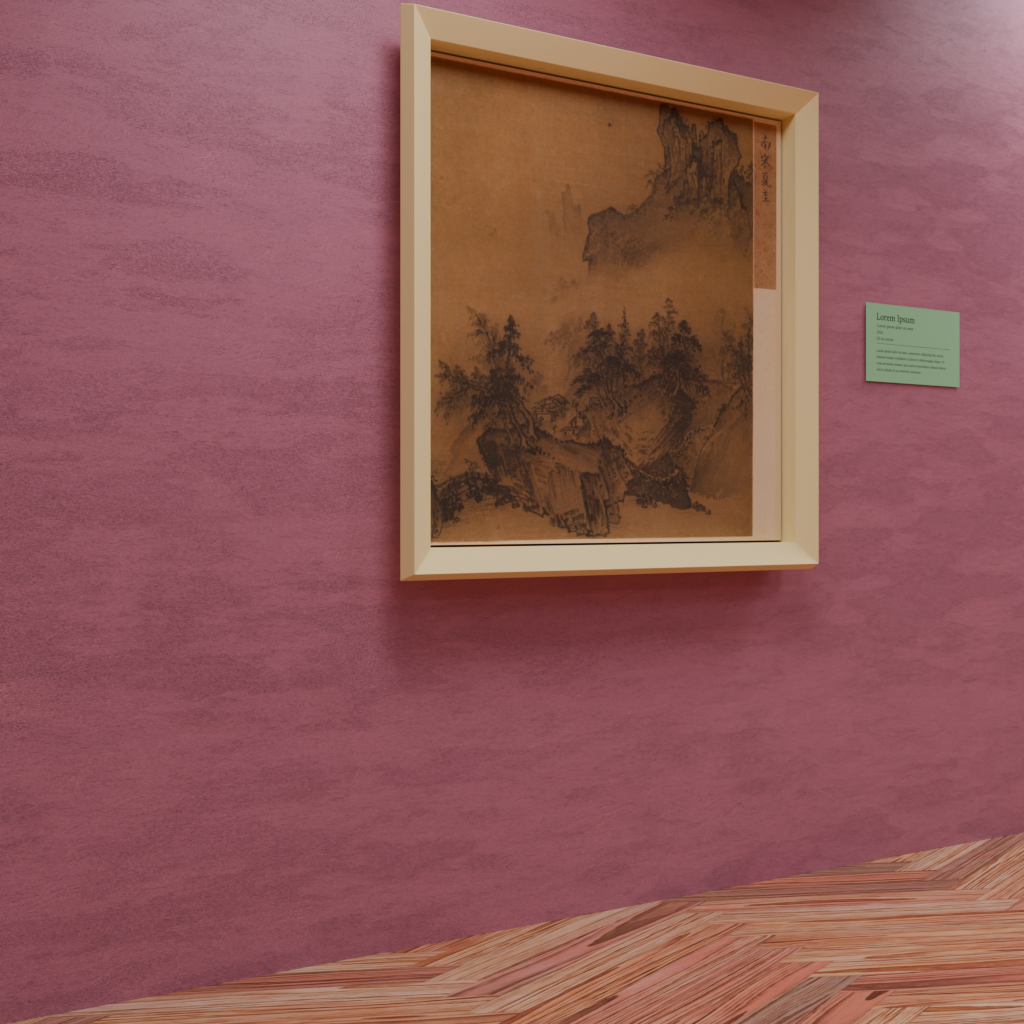} &
        \includegraphics[width=\linewidth]{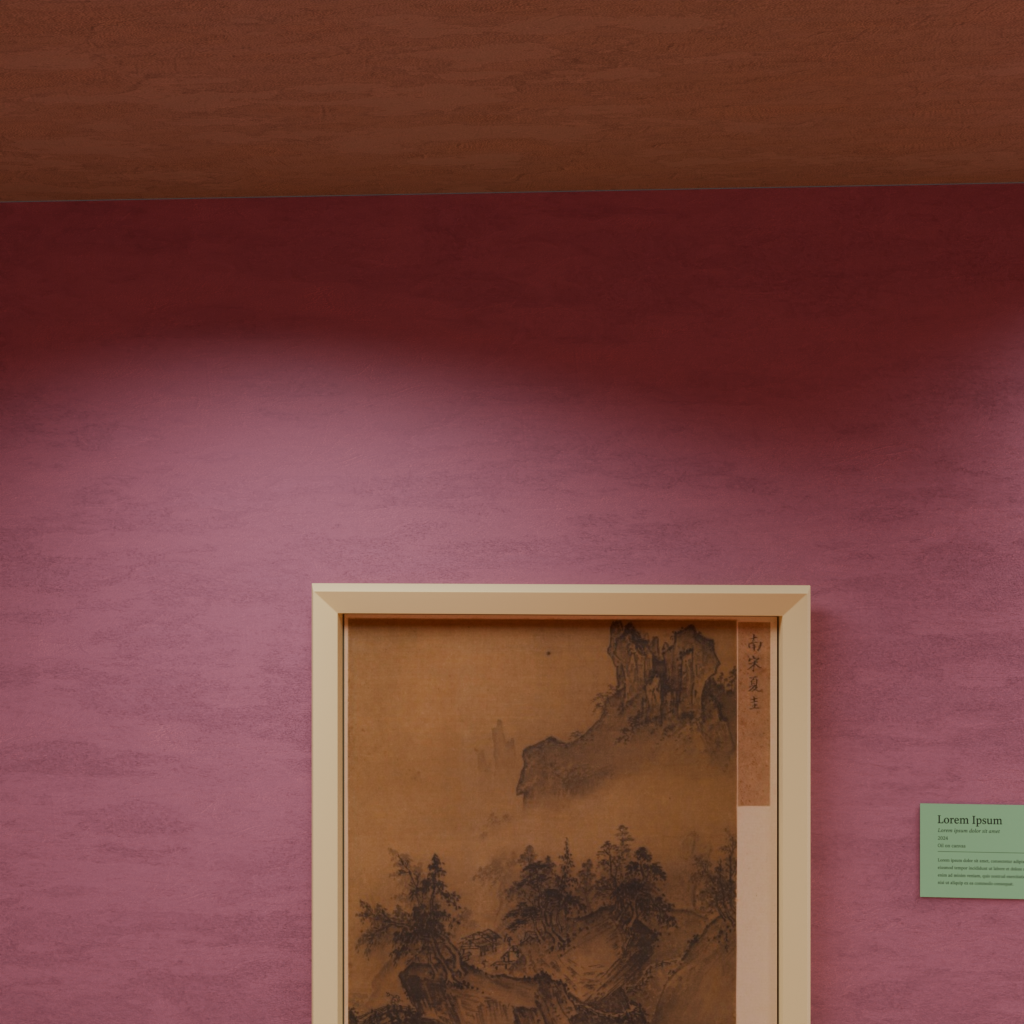} &
        \includegraphics[width=\linewidth]{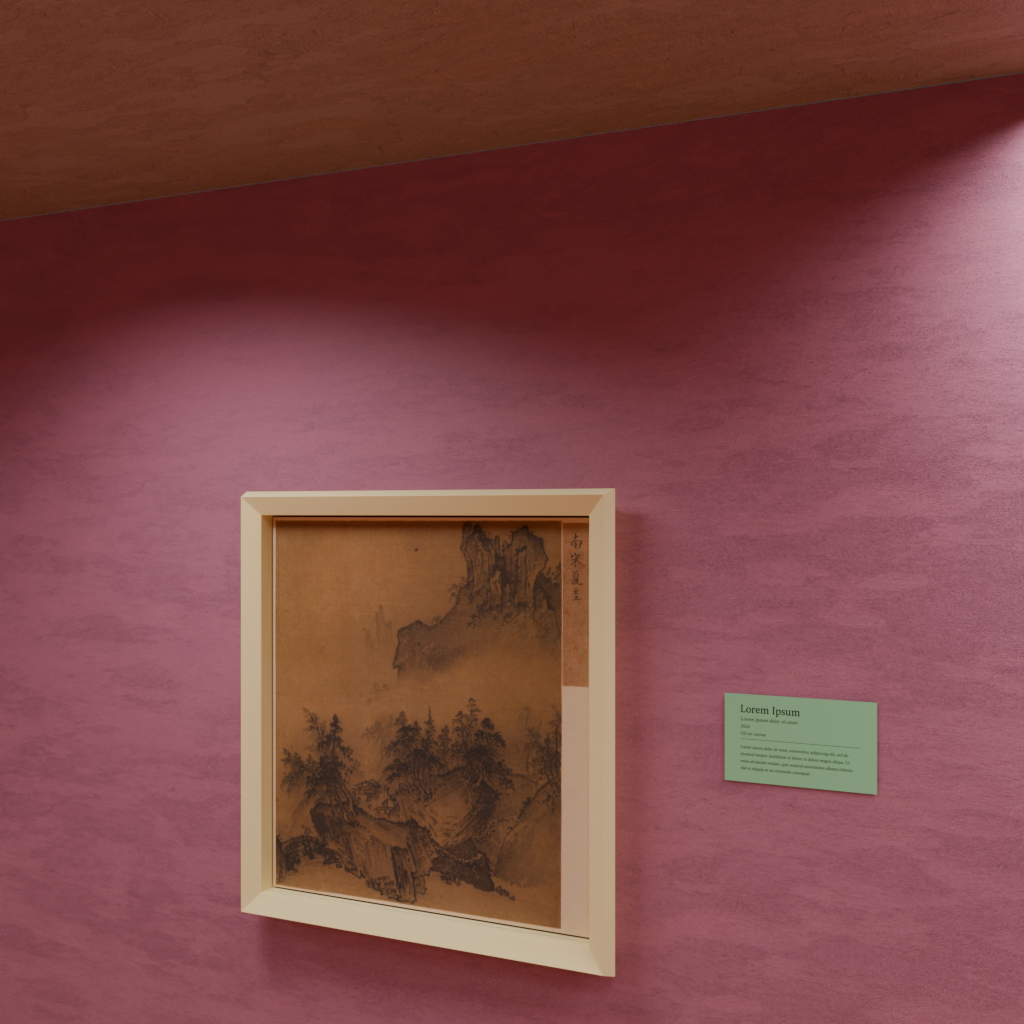} &
        \includegraphics[width=\linewidth]{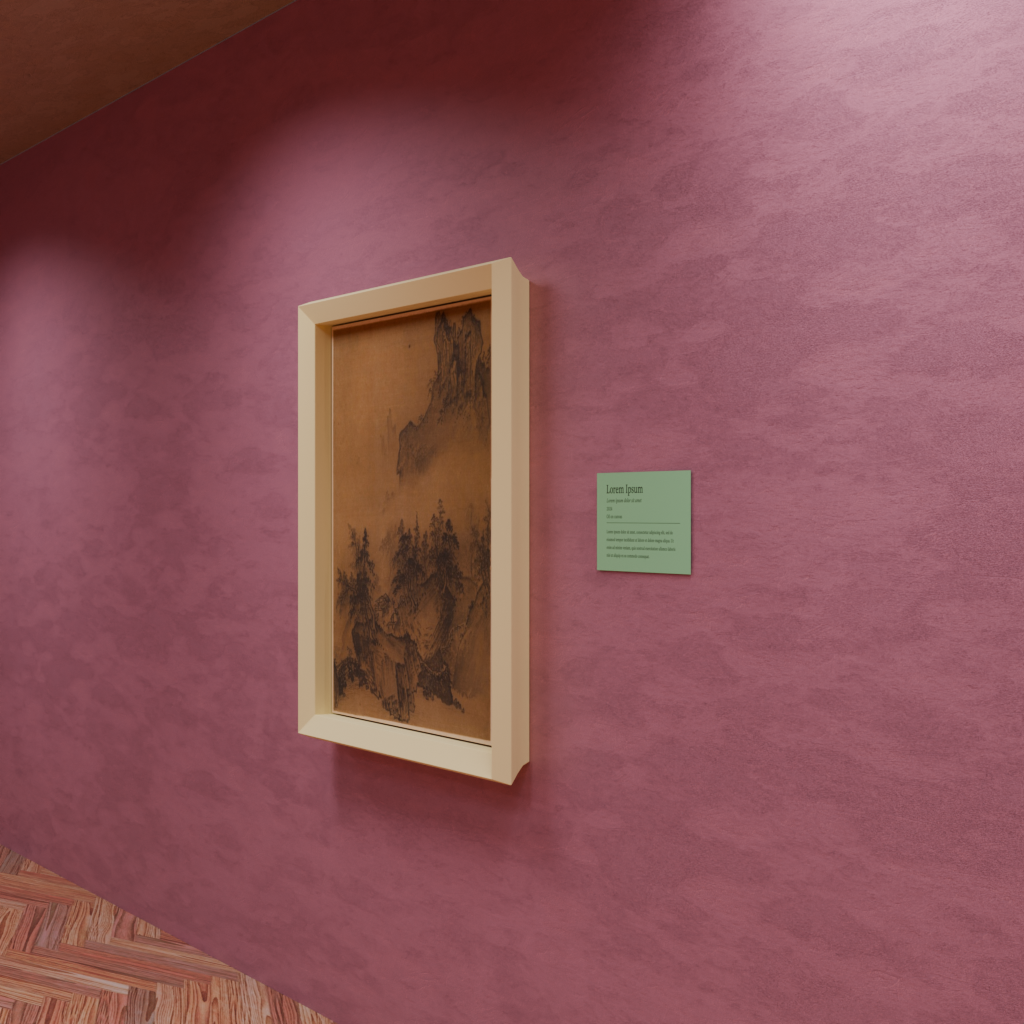} \\
        \includegraphics[width=\linewidth]{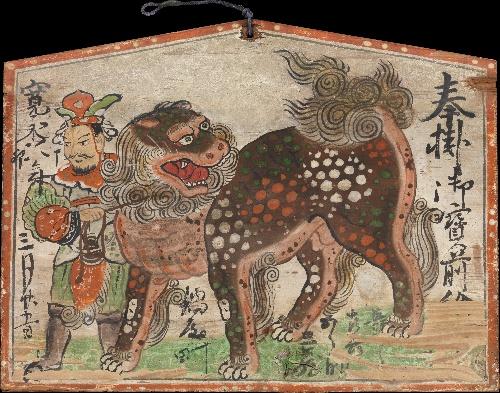} &
        \includegraphics[width=\linewidth]{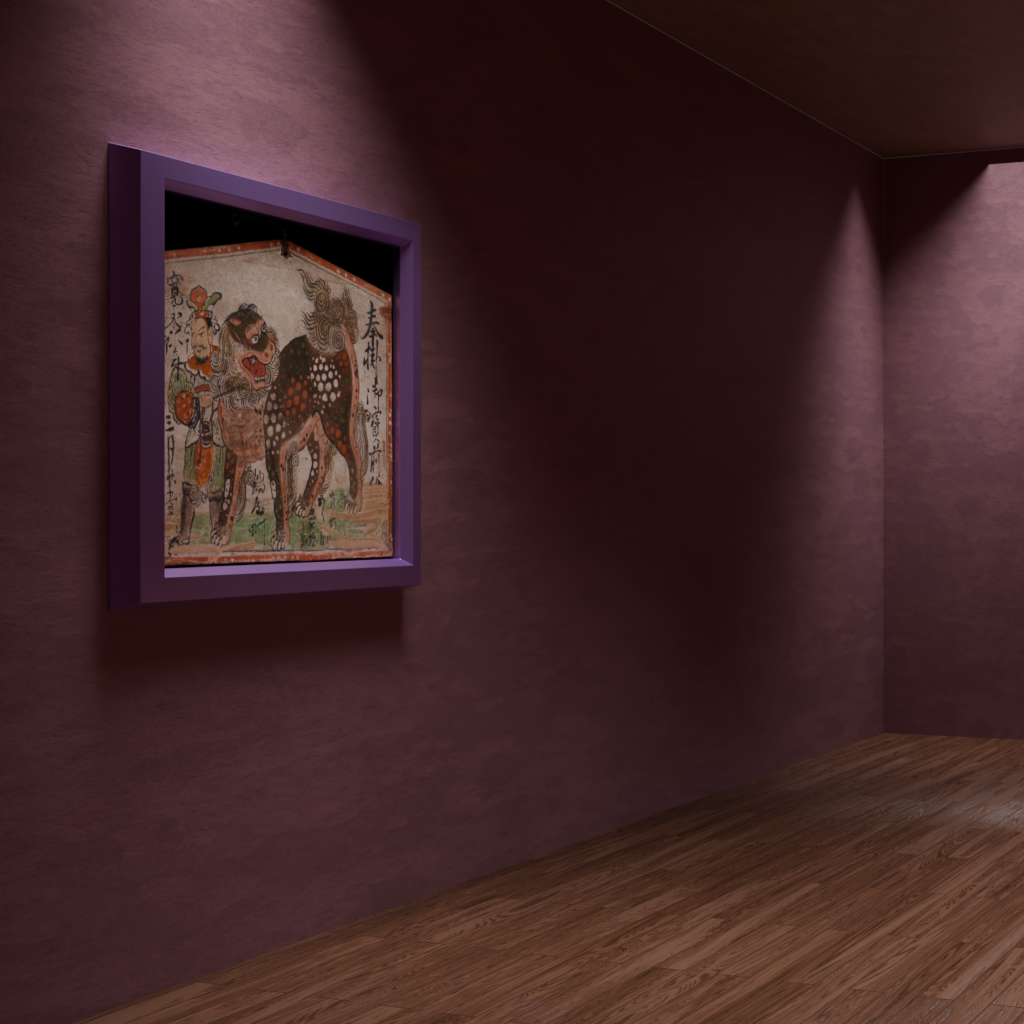} &
        \includegraphics[width=\linewidth]{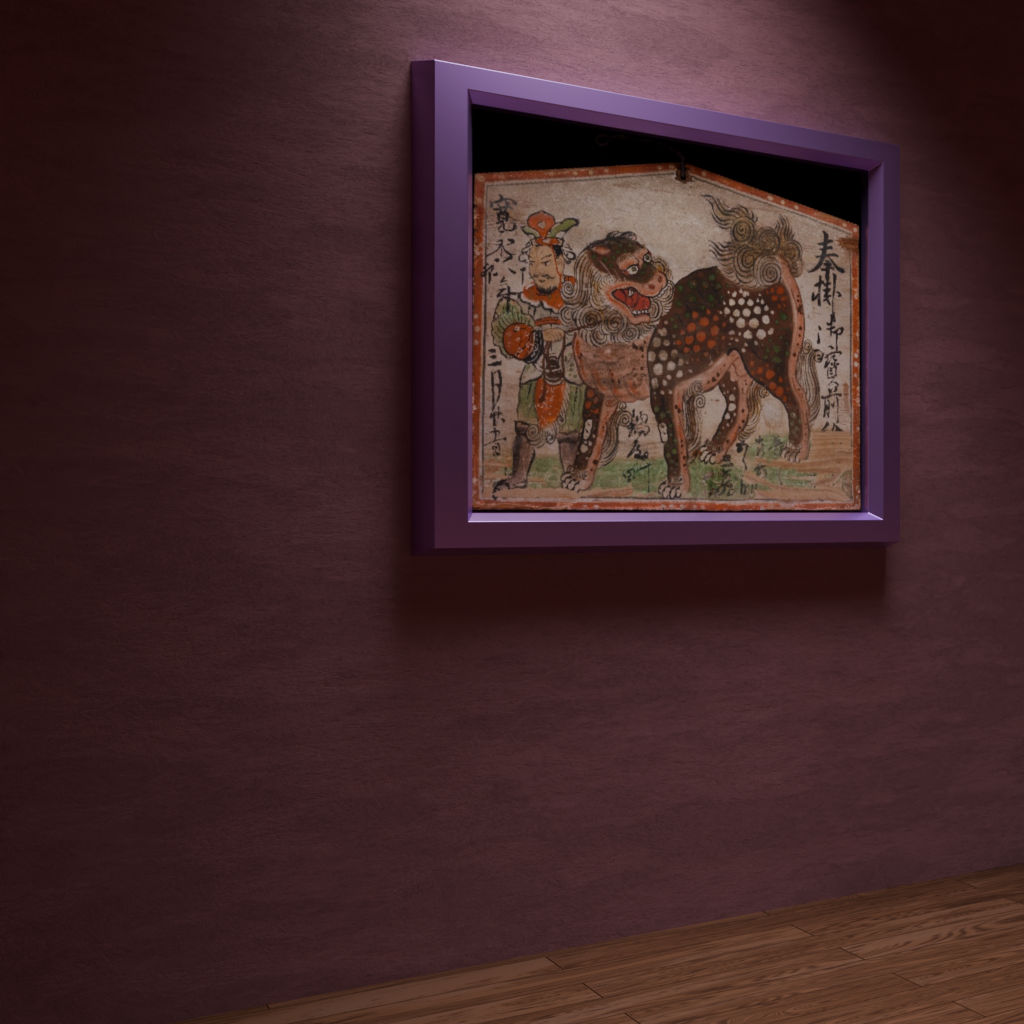} &
        \includegraphics[width=\linewidth]{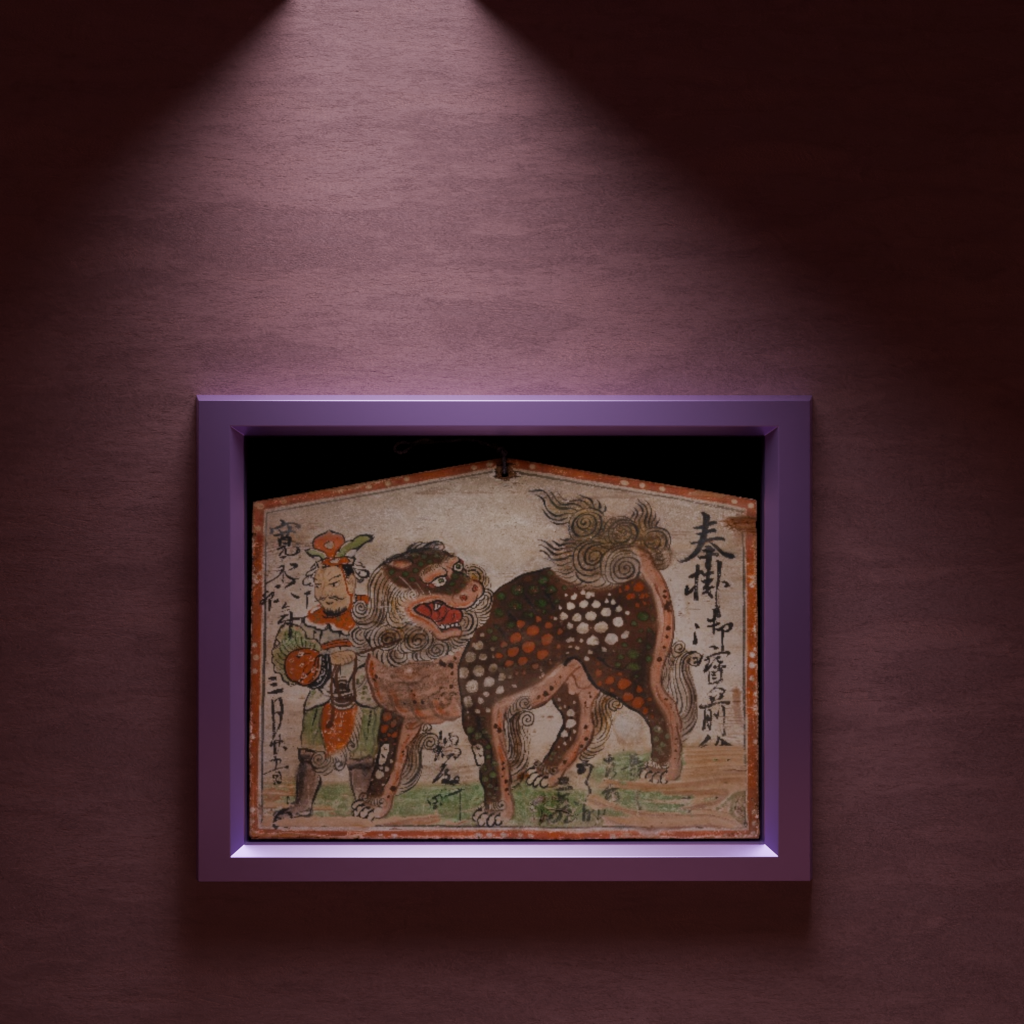} &
        \includegraphics[width=\linewidth]{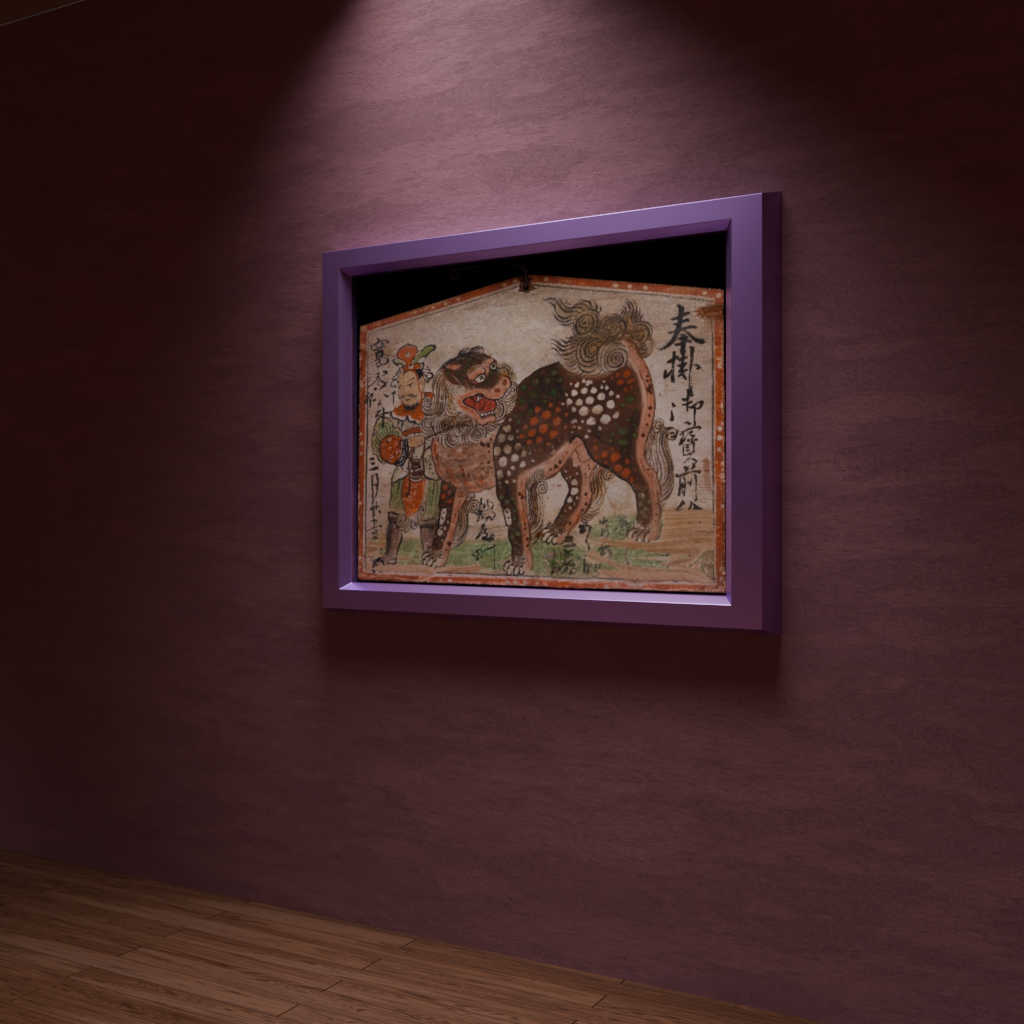} &
        \includegraphics[width=\linewidth]{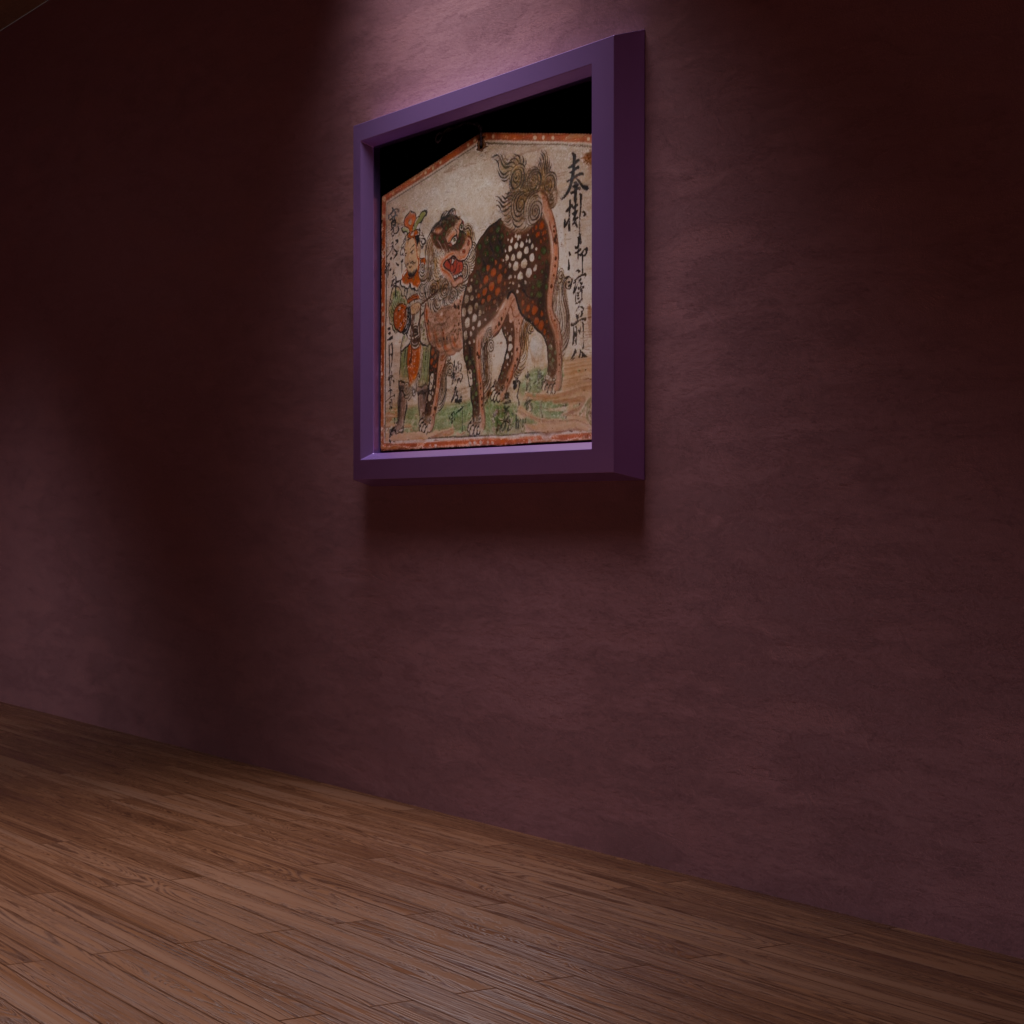} \\
    \end{tabular}
    \caption{Met source images (left) and the \syngallery{} renders generated from them. The catalog image is used directly as the canvas texture, so each artwork's identity is preserved exactly while the surrounding gallery scene is randomized.}
    \label{fig:met_syngal_comp}
\end{figure}

\begin{figure}[tbp]
\centering
\includegraphics[width=0.9\linewidth]{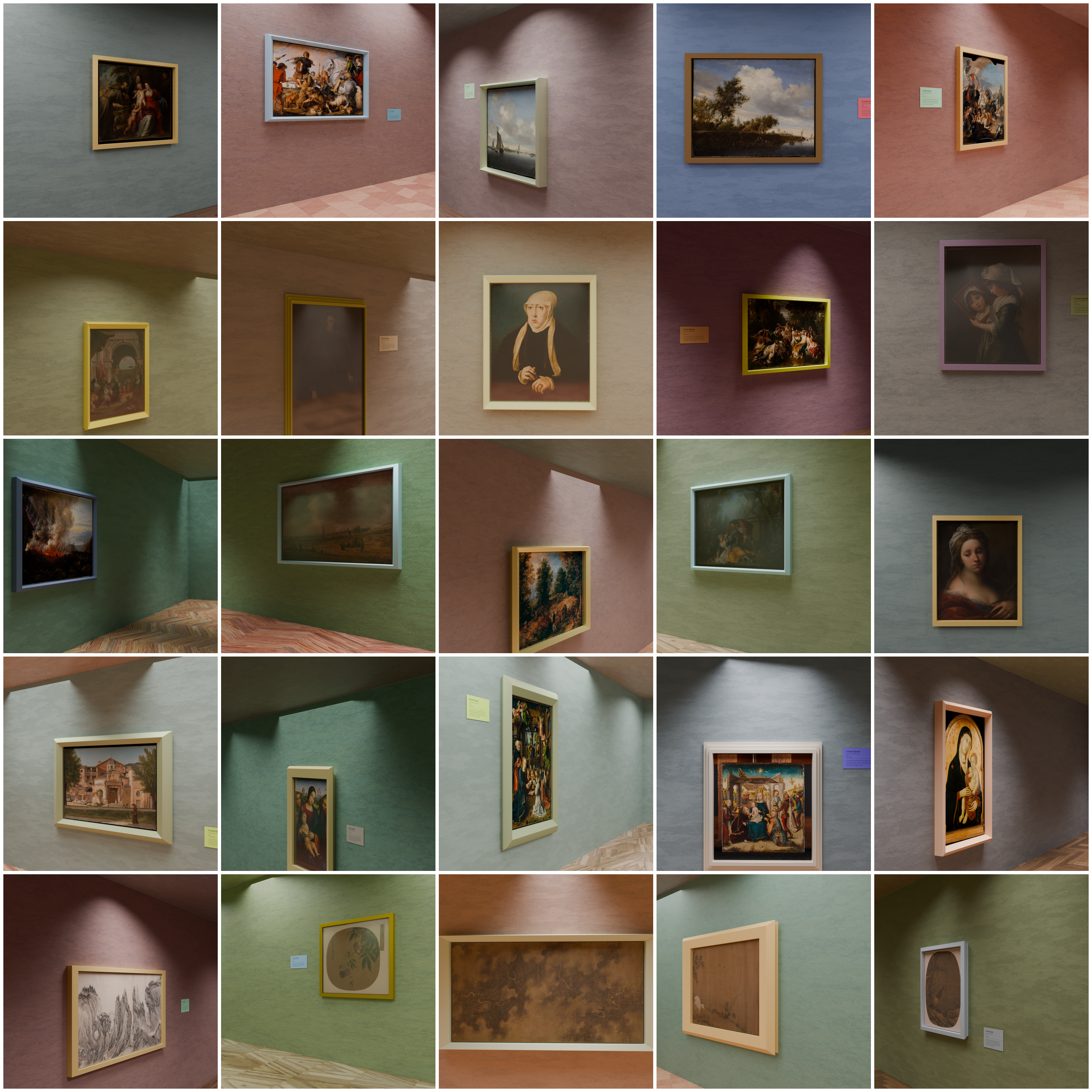}
\caption{An uncurated random sample of \syngallery{} renders, illustrating the
variance of the procedural data generation.}
\label{fig:variations}
\end{figure}

An overview of our framework is shown in Fig.~\ref{fig:overview}. Each 3D scene is composed of a rectangular room with a textured canvas placed on one wall, a frame, optional glass, an optional plaque, ceiling lights, and five fixed cameras placed at $30^\circ$, $60^\circ$, $90^\circ$, $120^\circ$, and $150^\circ$ around the painting (examples shown in Fig.~\ref{fig:met_syngal_comp}). The Met catalog image is applied as the canvas texture and the canvas is scaled to the image's aspect ratio, so the source artwork itself is never cropped, retouched, or replaced: the texture on the canvas is the unmodified catalog image. The rendered views naturally undergo the perspective projection of each camera, so oblique viewpoints foreshorten and partially clip the visible painting (see Fig.~\ref{fig:met_syngal_comp}); this transformation is a property of the observation, not an edit of the underlying artwork, and it is exactly the geometric variation that visitor photographs exhibit. The five cameras lie on a horizontal arc at a fixed height and distance from the canvas, all aimed at the painting, so they pan around the painting rather than tilt: $90^\circ$ is frontal, while $30^\circ$ and $150^\circ$ are the extreme oblique views. In rig terms the five poses correspond to rotations of $60^\circ$, $30^\circ$, $0^\circ$, $-30^\circ$, and $-60^\circ$ about the vertical (global $Z$) axis. Every painting is rendered from all five cameras, giving $4{,}898\times5=24{,}490$ RGB images at $512^2$ resolution. The renderer, camera distance and field of view, and the default output resolution are listed in Tab.~\ref{tab:params}.  Fig.~\ref{fig:variations} shows additional uncurated samples illustrating the variance of the procedural generation.

\subsection{Scene Randomization}

For every painting we sample one gallery configuration and render it from the selected cameras. Configuration randomizes the lighting (ceiling-light placement, intensity, and shape), the room appearance (wall, ceiling, and floor materials), the frame (geometry and material), the plaque (visibility, color, and placement), the presence of glass, and a small camera jitter. Tab.~\ref{tab:params} specifies, for each factor, the distribution it is drawn from and its range or set of values. The camera jitter is drawn from a uniform distribution, sampled once per view. Ceiling lights are positioned by Poisson-disk sampling, which enforces a minimum spacing. Glass is included as an independent Bernoulli event with probability $p=0.25$, and the frame geometry is drawn uniformly from a library of $27$ shapes. The source painting is loaded from the Met catalog and rescaled to preserve its aspect ratio.

\begin{table}[tbp]
% ---PART OF CAPTION: The last column indicates whether a factor is controlled in the ablation (Sec.~\ref{sec:results}): independently varied (\emph{yes}), varied only as part of a group such as lighting or room appearance (\emph{grouped}), or held fixed (\emph{no}).
  \caption{Generation parameters. Artwork identity is the only factor that is never randomized; the remaining factors change the viewing conditions under which the same painting is observed.}
  \label{tab:params}
  \centering\footnotesize
  \resizebox{\linewidth}{!}{%
  \begin{tabular}{l l p{0.5\linewidth}}
    \toprule
    Factor & Distribution & Range / values \\
    \midrule
    Viewpoint             & fixed                          & five angles $30^\circ,60^\circ,90^\circ,120^\circ,150^\circ$ about the vertical axis; optional choice of any subset \\
    Camera location / FOV & fixed       & location, field of view \\
    Camera jitter         & uniform                        & translation along $Z\in [-1;1]$; translation along $Y\in [-0.5;0.5]$; rotation about $Z\in [-5^{\circ};5^{\circ}]$; separate values for each camera \\
    Light placement       & Poisson-disk& min. spacing, max.\ density \\
    Light intensity       & uniform     & single value shared by all lights $\in [0.25;0.8]$ \\
    Light shape           & discrete, optional fixed; uniform     & $\{$square, disk$\}$, optional choice of base shape; rotation, width, depth\\
    Wall / ceiling color  & uniform     & separate values for each RGB channel $\in [0;1]$ \\
    Floor material        & discrete uniform               & over 5 materials \\
    Frame geometry        & discrete uniform               & over $27$ frame shapes \\
    Frame material        & uniform     & separate values for each RGB channel, roughness and metallic $\in [0;1]$ \\
    Plaque visibility     & Bernoulli   & present with probability $p=0.5$ \\
    Plaque color          & uniform     & separate values for each RGB channel $\in [0;1]$ \\
    Plaque placement      & discrete                       & offset relative to painting along $\pm$~X, random choice from a set of three values along Z \\
    Glass presence        & Bernoulli   & present with probability $p=0.25$ \\
    Renderer / samples    & fixed                          & sample count; engine (e.g.\ Cycles), render modes (e.g.\ RGB, semantic mask) \\
    Render resolution     & fixed                          & default $512^2$; high-resolution variant $1024^2$ \\
    Artwork identity      & deterministic                  & Met image as canvas texture, aspect ratio preserved \\
    \bottomrule
  \end{tabular}%
  }
\end{table}

% \subsection{Dataset variants}

% The full five-view dataset contains $24{,}490$ images; we also provide a three-view subset ($14{,}694$) and a frontal-only subset ($4{,}898$, the $90^\circ$ view), together with frozen-room, frozen-frame, and high-resolution ($1024^2$) variants, where the frozen variants hold the corresponding factor of Tab.~\ref{tab:params} fixed. For each painting the generator samples a gallery configuration, maps the catalog image to the canvas, applies the shared camera jitter, and renders the five RGB views, producing the ablation variants by freezing selected scene factors or restricting the camera set. \syngallery{} contains RGB renders only and carries no depth, mask, or camera-parameter annotations.

\section{Evaluation Protocol}
\label{sec:evaluation}

\subsection{The Met Benchmark}
\label{sec:met_benchmark}

All experiments are conducted on the Met dataset, an instance-level recognition benchmark built on the open-access collection of the Metropolitan Museum of Art. Its training set contains 397{,}121 catalog images covering 224{,}408 exhibit classes, photographed under controlled studio conditions and extremely long-tailed: each exhibit has at most ten catalog images, and 60.8\% of the classes have only one.

The realistic museum views that make the benchmark difficult are supplied entirely on the query side. Met queries are real photographs taken on site by museum visitors (39 in total), partly collected by the benchmark team and partly gathered from public Flickr groups; they exhibit exactly the scene-level conditions the catalog lacks, such as oblique viewpoints, gallery illumination, frames, glass, and reflections. The test set combines 1{,}003 such Met queries with 18{,}316 distractor queries crawled from Wikimedia Commons: 10{,}352 \emph{other-artwork} queries depicting artworks outside the Met collection and 7{,}964 \emph{non-artwork} queries depicting generic content. A validation set (129 Met queries, 2{,}036 distractors) is reserved for hyper-parameter tuning. Every query is matched against the full 397k-image training corpus, which also serves as the retrieval database; distractors have no correct match and must receive lower prediction confidence than the Met queries.

We evaluate in two settings. The \emph{full benchmark} follows the protocol above. The \emph{closed painting world} restricts the task to the painting subset from which \syngallery{} is generated: the 12{,}403 catalog photographs of the 4{,}898 painting classes form the database, and the 148 real visitor photographs of these paintings in the test set form the queries. This setting contains no distractors and therefore isolates painting recognition from out-of-distribution rejection.

\subsection{Model Architecture and Training}
To ensure a direct and fair comparison, we reproduce the architecture of the best single model from the original Met benchmark. We use a ResNet-18 backbone~\cite{resnet} pre-trained using Semi-Weakly Supervised Learning (SWSL)~\cite{swsl}. The model is trained via a contrastive loss~\cite{constrastive} with generalized-mean (GeM) pooling and a whitening fully-connected layer. During training with synthetic data, positive pairs are formed by mining the closest samples across both the real and synthetic domains. At inference time, we extract multi-scale descriptors to compute the cosine similarity scores.

\subsection{Metrics}

We evaluate our models using the open-set retrieval protocol of the Met benchmark. Performance is measured using average classification accuracy (ACC) and Global Average Precision (GAP), two standard metrics for instance-level recognition. Following the benchmark, classification is performed using a $k$-nearest-neighbor ($k$NN) classifier operating on image embeddings. Given a query image $q$ and a database image $x$, their similarity is measured using the cosine score $v(x)^\top v(q)$. The confidence assigned to class $c$ is

\begin{equation}
\label{eq:class-confidence}
s_c(q)
=
\max_{\substack{x \in NN_k(q)}}
v(x)^\top v(q),
\end{equation}

where $NN_k(q)$ denotes the set of the $k$ nearest neighbors of $q$ in embedding space. The resulting class confidences are normalized using a temperature-scaled softmax (with temperature coefficient $\tau$).

Retrieval quality is measured by ranking all queries according to their normalized prediction confidence. GAP is computed as $\mathrm{GAP}=\frac{1}{M}\sum_{i=1}^{T}p(i)r(i)$, where $p(i)$ is the precision at rank $i$, $r(i)$ indicates whether the prediction at position $i$ is correct, $M$ is the number of target Met queries, and $T$ is the total number of evaluated queries. We report two variants of this metric. $\GAPm$ is evaluated only on target artwork queries and therefore measures closed-world painting recognition. $\GAP$ additionally includes the out-of-distribution distractor queries described in Sec.~\ref{sec:met_benchmark} and corresponds to the full open-set retrieval setting of the benchmark. Since distractor queries are always counted as incorrect predictions, high $\GAP$ requires both accurate artwork recognition and reliable confidence calibration.

The retrieval hyper-parameters $k$ and $\tau$ are selected using the validation protocol of the Met dataset. 
%For the full benchmark, they are tuned on the validation set following the Met protocol.
For the ablation experiments, we use 2-fold cross-validation on the painting queries.
An audit of this tuning procedure is provided in Appendix~\ref{sec:appendix_honest_tuning}.

 \section{Results}
  \label{sec:results}

  \subsection{Performance on the Met Benchmark}
\begin{figure}[tbp]
\centering
\includegraphics[width=0.62\linewidth]{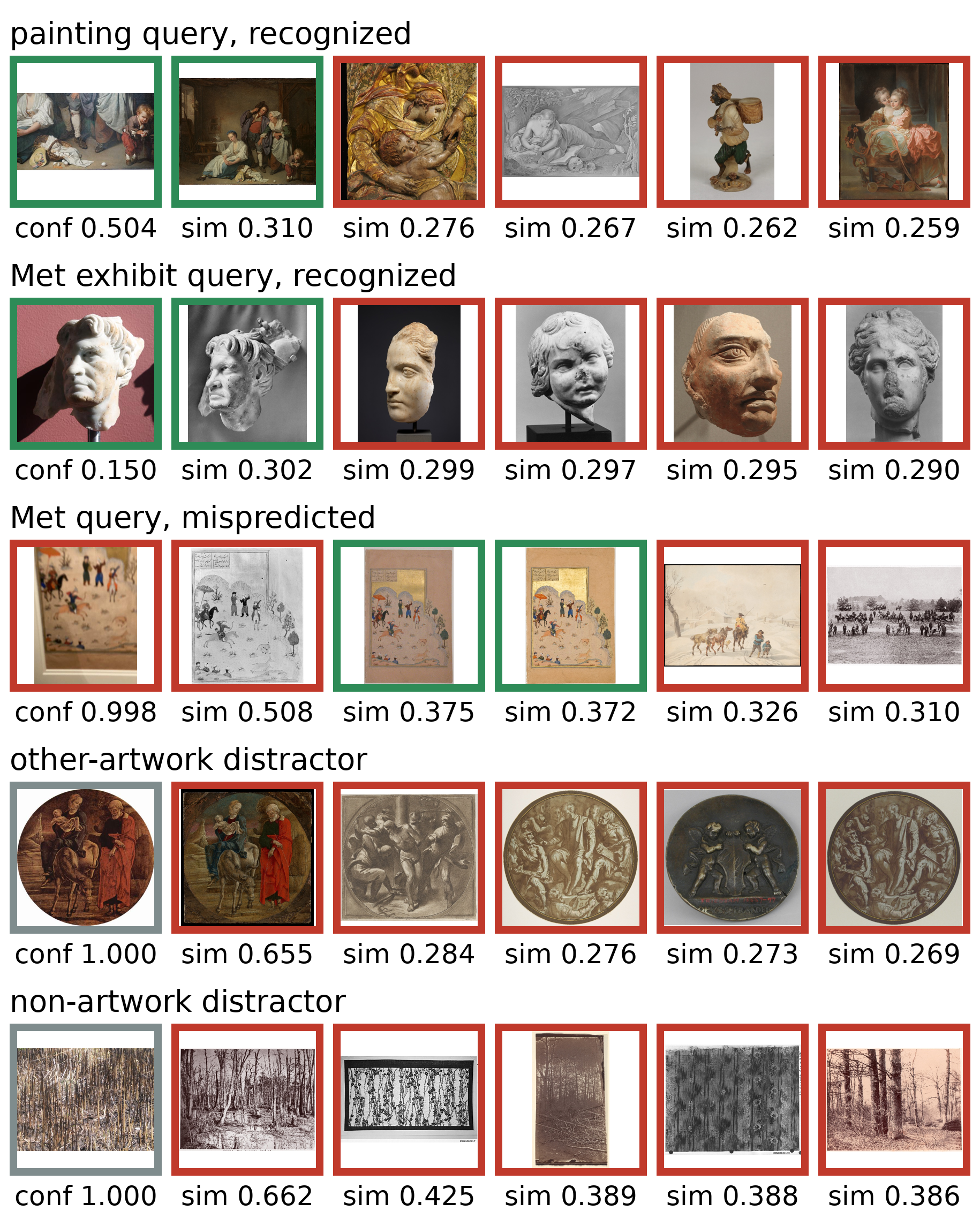}
\caption{\textbf{Retrieval in the full open-set benchmark} with our best model.
Each row shows a query (left, with its kNN-softmax confidence) and its five
nearest database images (with cosine similarity): a painting query, a
non-painting Met exhibit query, a mispredicted Met query, and the two
highest-confidence distractor queries. Green frames mark database images of the
query's class, red frames all others; the query border encodes the outcome
(green recognized, red mispredicted, gray distractor). The mispredicted query
is confused with a near-identical folio of a different manuscript page, and the
distractors retrieve visually similar Met exhibits with maximal confidence,
which the $\GAP$ metric penalizes.}
\label{fig:qualitative}
\end{figure}

  We first evaluate whether \syngallery{} renders improve retrieval under the full Met
  benchmark protocol. As shown in Tab.~\ref{tab:main}, adding \syngallery{} to the same
  training recipe increases full-benchmark $\GAP$ from 35.97 to 38.48. This provides a
  direct comparison in which the only change is the addition of synthetic gallery views,
  yielding a $+2.51$ $\GAP$ improvement. The result also exceeds the best published single
  model under the same protocol ($\GAP$ 36.1), without changing the retrieval architecture or adding external real-world pre-training data. Fine-tuning the epoch-10 baseline on the synthetic renders alone gives the highest $\GAP$
  of 38.99. The corresponding non-distractor metrics also improve over the real-only
  reproduction ($\GAPm$ 55.15, ACC 57.23). This indicates that the renders provide useful
  invariances to gallery-specific variation, such as viewpoint and illumination, while
  preserving the instance identity needed for retrieval.
  Fig.~\ref{fig:qualitative} illustrates the difficulty of the full open-set
  setting: handheld Met queries must be matched against 397k catalog images,
  while every distractor query must receive lower confidence than the correctly
  recognized Met queries.

  % --- Table: main full-benchmark results
  \begin{table}[tbp]
  \caption{Full Met benchmark: 1{,}003 Met queries plus 18{,}316 distractors against the
  397k-image database. Adding SynGallery to the same training set beats the best published
  single model under its exact setup.}
  \label{tab:main}
  \centering\footnotesize
  \scalebox{0.9}{
  \begin{tabular}{lccc}  
  \toprule
  model (training data) & $\GAP$ & $\GAPm$ & ACC \\
  \midrule
  best published single model & 36.1 & 52.4 & 55.0 \\
  our reproduction (397k real) & 35.97 & 52.14 & 54.64 \\
  \midrule
  from scratch, 397k real $+$ SynGallery & 38.48 & 55.04 & \textbf{57.63} \\
  fine-tuned, $+$ SynGallery only & \textbf{38.99} & \textbf{55.15} & 57.23 \\
  fine-tuned, $+$ SynGallery $+$ real & 38.66 & 53.73 & 55.53 \\
  \bottomrule
  \end{tabular}
  }
  \end{table}

  \subsection{Real vs. Synthetic Mixing and Scaling}
\begin{figure}[tbp]
  \centering
  \includegraphics[width=\linewidth]{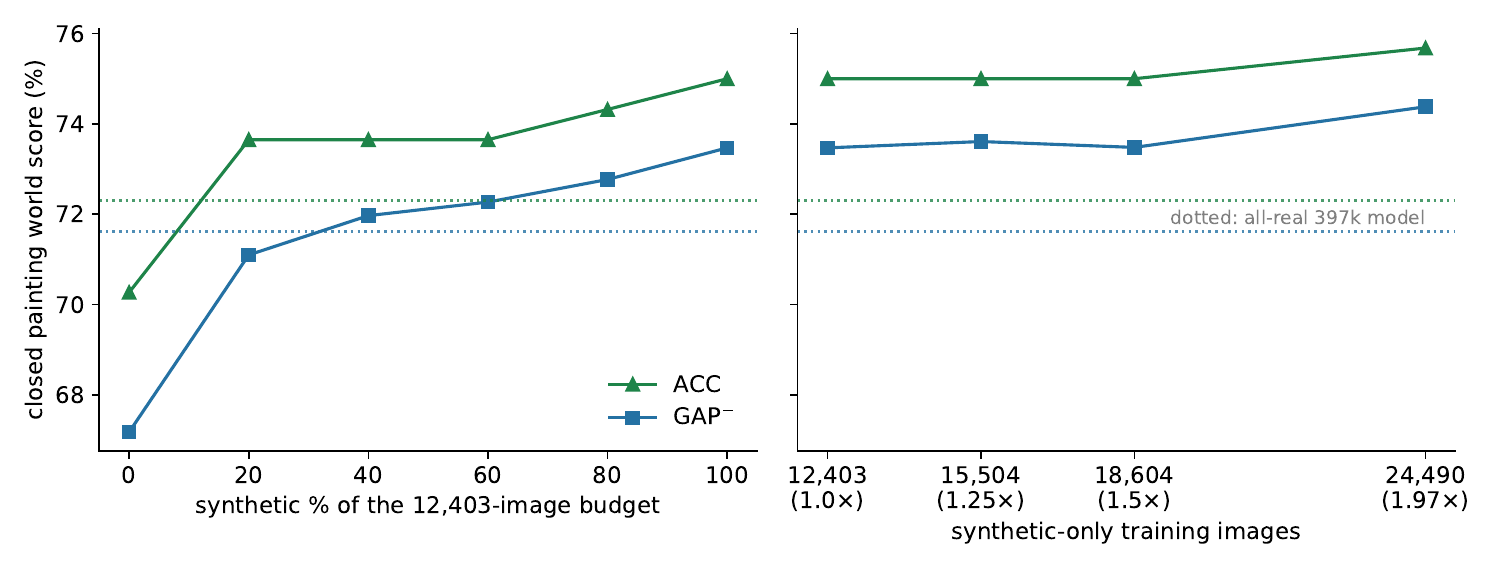}
  \caption{\textbf{Performance on the full Met benchmark} (1{,}003 Met queries, 18{,}316 distractors, 397k-image database). Adding the SynGallery renders to the baseline training set improves upon the best published single model under the exact same protocol ($\GAP$ 38.48 vs.\ 36.1).}
  \label{fig:mix}
  \end{figure}

  To isolate the contribution of the synthetic images relative to the catalog photographs
  from which they are derived, we evaluate real-to-synthetic training mixtures at a fixed
  budget of 12{,}403 images. This budget corresponds exactly to the complete set of real studio photographs available in the full 397k database for the 4{,}898 painting classes used to generate \syngallery{}. As shown in Tab.~\ref{tab:mix} and Fig.~\ref{fig:mix},
  replacing real studio photographs with synthetic renders consistently improves
  performance. The all-synthetic mix achieves a closed-world $\GAPm$ of 73.47, compared to
  67.18 for the all-real baseline, showing that multi-view gallery renders provide a
  stronger training signal than the original studio views at the same image budget.
%
% --- Figure: mix/scaling exact values
  Increasing the amount of synthetic-only training data beyond the fixed budget, up to the
  full set of 24{,}490 renders, gives further gains. At this scale, the synthetic-only
  model reaches $\GAPm$ 53.78 and ACC 56.63 on the full benchmark. It therefore exceeds the
  397k real-image baseline on both non-distractor metrics, using approximately $16\times$
  fewer training images. Its open-set $\GAP$ remains
  lower (34.38 vs.\ 35.97), which is expected since the synthetic-only model does not
  include distractor-class training data such as photographs of busts or statues.

  % --- Table: mix/scaling exact values
  \begin{table}[tbp]
  \caption{\textbf{Real-to-synthetic data mixing and scaling.} At a strictly matched
12{,}403-image budget, the all-synthetic mix uniformly outperforms the all-real
baseline; scaling to all 24{,}490 renders surpasses the 397k all-real model on both
$\GAPm$ and ACC.}
  \label{tab:mix}
  \centering\footnotesize
  \scalebox{0.9}{
  \begin{tabular}{lccccc}
  \toprule
  & \multicolumn{2}{c}{closed painting world} & \multicolumn{3}{c}{full benchmark} \\
  \cmidrule(lr){2-3}\cmidrule(lr){4-6}
  training data & $\GAPm$ & ACC & $\GAP$ & $\GAPm$ & ACC \\
  \midrule
  100:0 --- all real & 67.18 & 70.27 & 28.83 & 49.08 & 52.14 \\
  80:20 & 71.10 & 73.65 & 30.33 & 50.41 & 53.44 \\
  60:40 & 71.97 & 73.65 & 31.23 & 50.84 & 53.54 \\
  40:60 & 72.27 & 73.65 & 31.26 & 51.12 & 53.64 \\
  20:80 & 72.77 & 74.32 & 32.29 & 51.85 & 54.34 \\
  0:100 --- all synthetic & 73.47 & 75.00 & 32.75 & 52.37 & 54.94 \\
  \midrule
  synth-only, 15{,}504 (1.25$\times$) & 73.61 & 75.00 & 33.26 & 52.78 & 55.23 \\
  synth-only, 18{,}604 (1.5$\times$) & 73.48 & 75.00 & 33.62 & 53.18 & 55.73 \\
  synth-only, all 24{,}490 (1.97$\times$) & \textbf{74.38} & \textbf{75.68} & 34.38 &
  \textbf{53.78} & \textbf{56.63} \\
  \midrule
  all-real reference (397k) & 71.62 & 72.30 & \textbf{35.97} & 52.14 & 54.64 \\
  \bottomrule
  \end{tabular}
  }
  \end{table}

\subsection{Embedding-Space Coverage Across Domains}
\label{sec:embedding_domains}

\begin{figure}[tbp]
\centering
\includegraphics[width=0.68\linewidth]{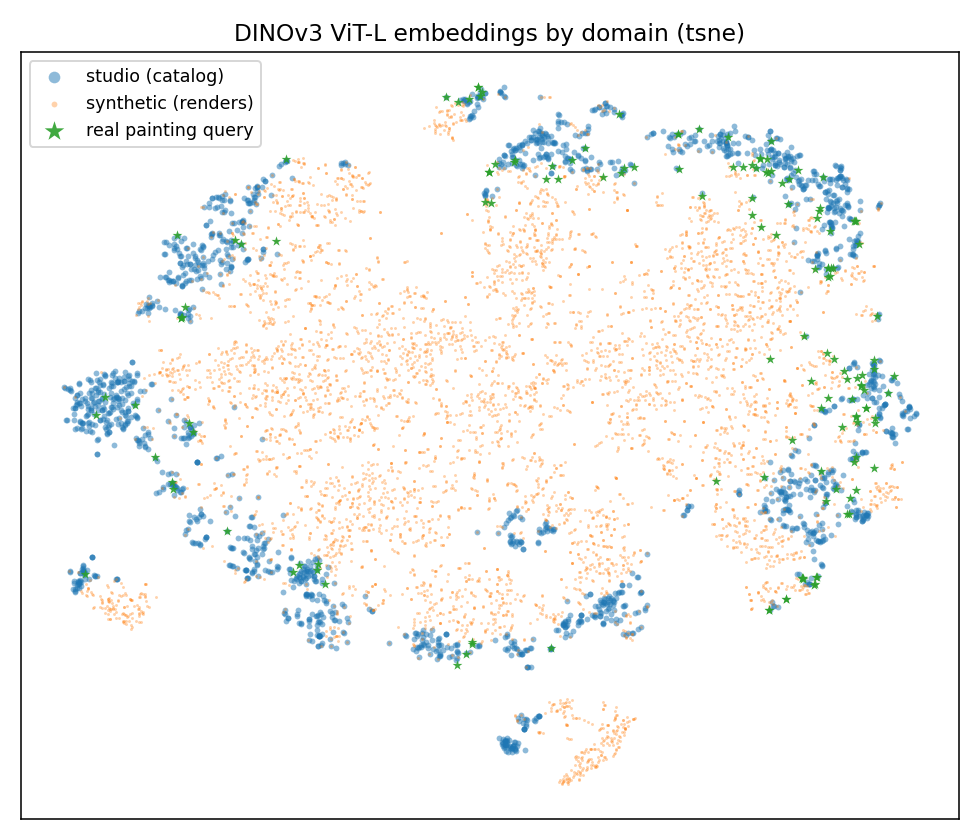}
\caption{\textbf{Embedding-space distribution across image domains.}
Two-dimensional t-SNE projection of DINOv3 ViT-L embeddings for studio
catalog photographs, synthetic \syngallery{} renders, and real painting
queries. Catalog photographs form compact local clusters, whereas the
synthetic renders span a broader and more continuous region of the
projected representation space. Real painting queries frequently occur
near catalog-image clusters and within regions also covered by the
synthetic renders.}
\label{fig:tsne}
\end{figure}

To qualitatively examine the relationship between the catalog photographs, synthetic renders, and real painting queries, we extract DINOv3 ViT-L embeddings and project them into two dimensions using t-SNE. As shown in Fig.~\ref{fig:tsne}, the three domains exhibit distinct but partially overlapping distributions.

The catalog photographs form relatively compact local clusters, many of which lie near the boundary of the projection. Real painting queries frequently occur within or close to these catalog-image neighborhoods, indicating that the pretrained representation retains correspondence between controlled catalog photographs and real photographs of paintings. In contrast, the synthetic renders occupy a broader and more continuous region of the embedding space. They extend beyond the compact catalog clusters and populate many of the intermediate regions between them.

This distribution is consistent with the intended role of \syngallery{}. The renders preserve artwork identity while expanding the range of appearances induced by viewpoint, framing, illumination, and gallery context. Rather than reproducing only the compact distribution of the source catalog images, the synthetic data provide broader representation-space coverage, including regions close to many real painting queries. This offers a qualitative explanation for the performance improvements observed when synthetic renders supplement or replace catalog photographs during training.

%The visualization should nevertheless be interpreted cautiously. t-SNE emphasizes local neighborhood structure and does not preserve global distances, relative cluster sizes, or density. Fig.~\ref{fig:tsne} therefore provides qualitative evidence of cross-domain coverage rather than a quantitative measure of domain alignment.

  \subsection{Dataset Ablation: The Mechanisms of Transfer}
\begin{figure}[tbp]
    \centering
    \setlength{\tabcolsep}{2pt}
    \begin{tabular}{NNNNNN}
        fixed scene & + textures & + light & + glass & + frame & + camera jitter \\
        \includegraphics[width=\linewidth]{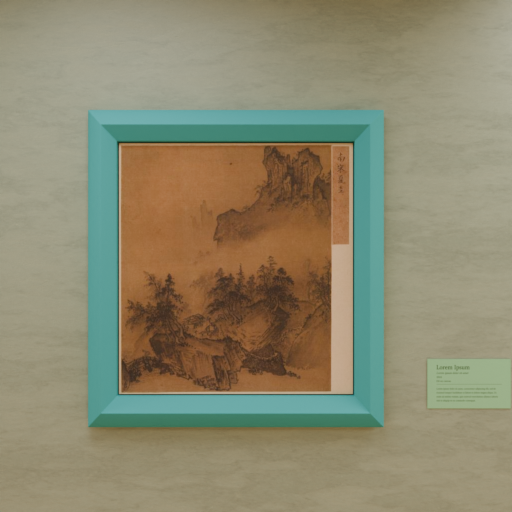} &
        \includegraphics[width=\linewidth]{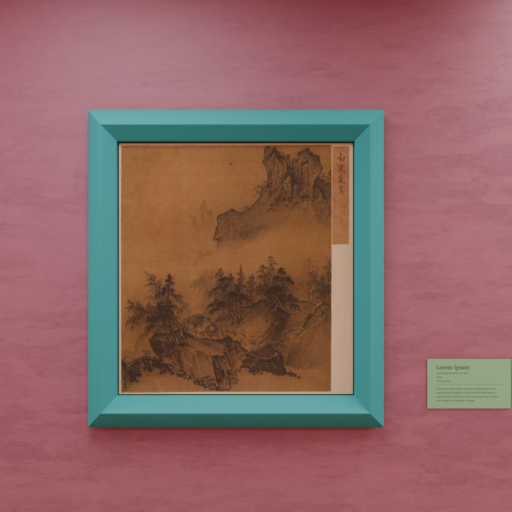} &
        \includegraphics[width=\linewidth]{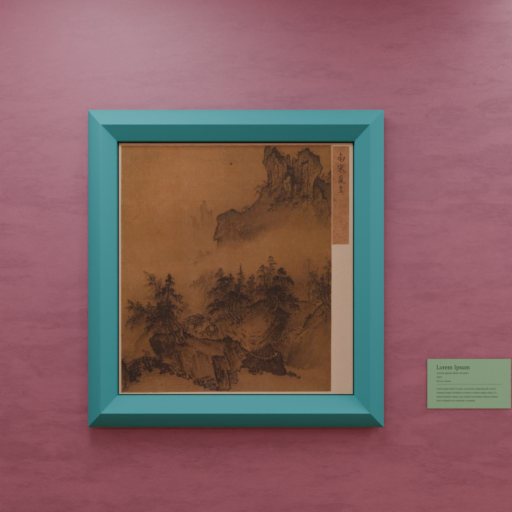} &
        \includegraphics[width=\linewidth]{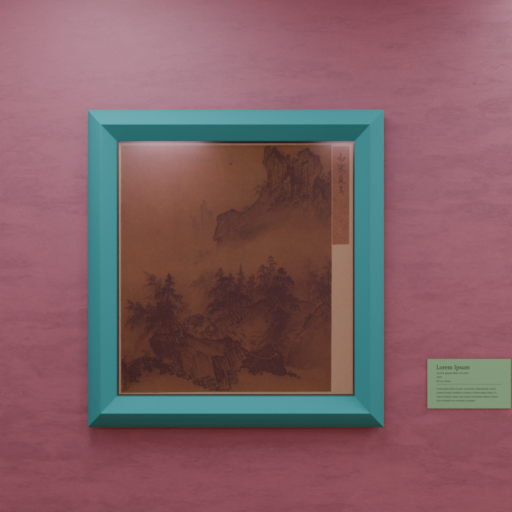} &
        \includegraphics[width=\linewidth]{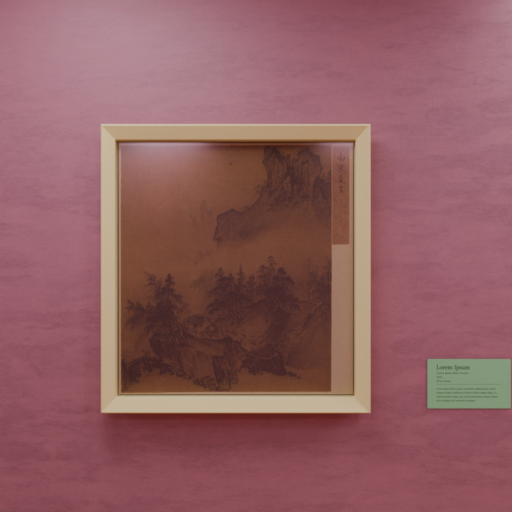} &
        \includegraphics[width=\linewidth]{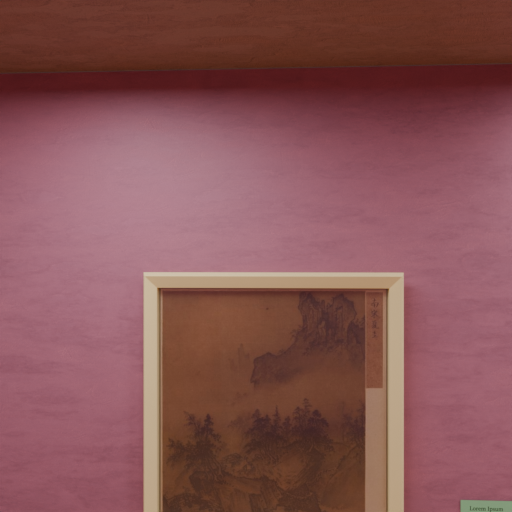} \\
        \includegraphics[width=\linewidth]{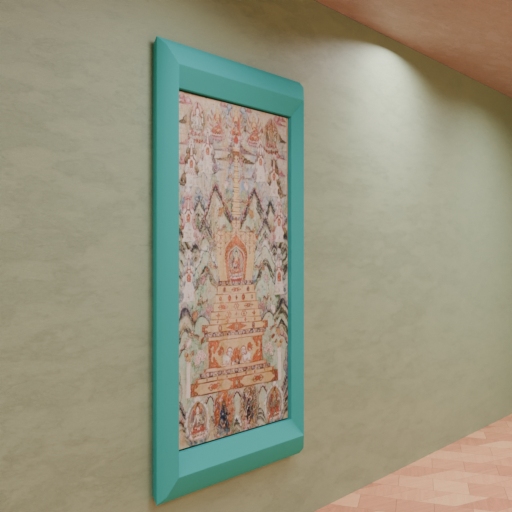} &
        \includegraphics[width=\linewidth]{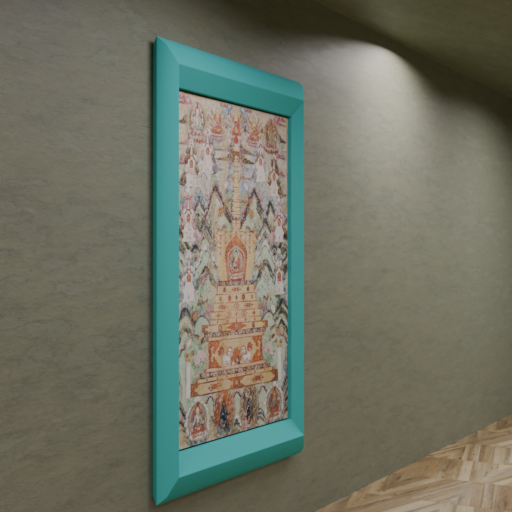} &
        \includegraphics[width=\linewidth]{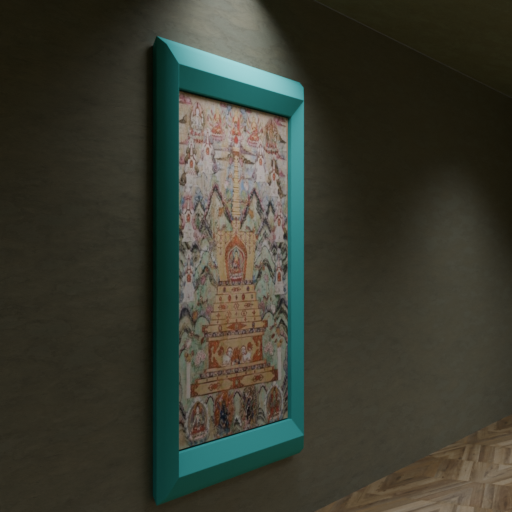} &
        \includegraphics[width=\linewidth]{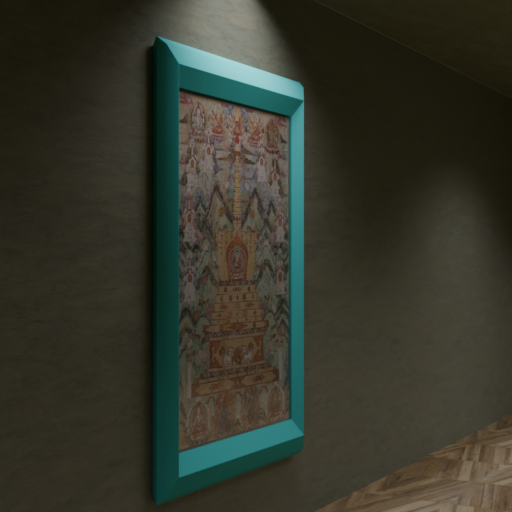} &
        \includegraphics[width=\linewidth]{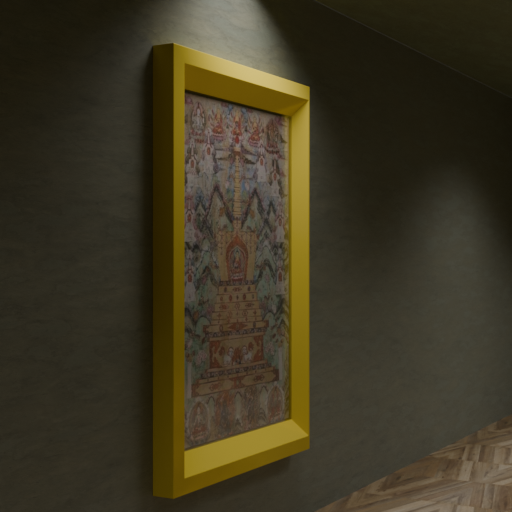} &
        \includegraphics[width=\linewidth]{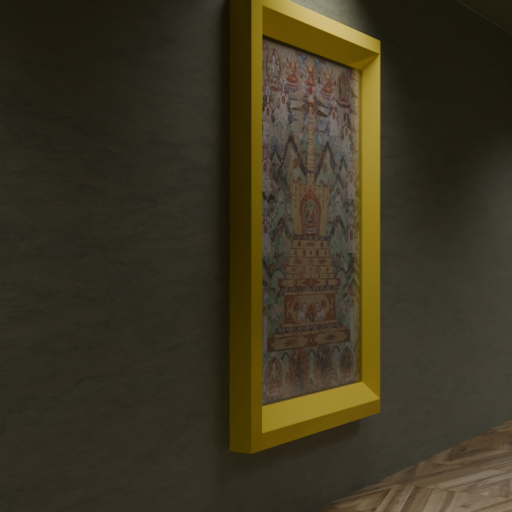} \\
        \includegraphics[width=\linewidth]{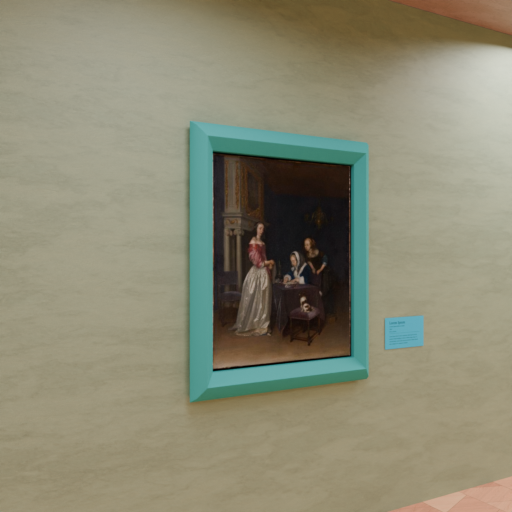} &
        \includegraphics[width=\linewidth]{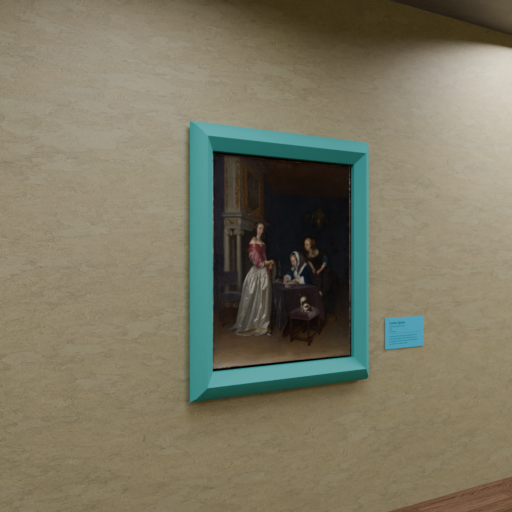} &
        \includegraphics[width=\linewidth]{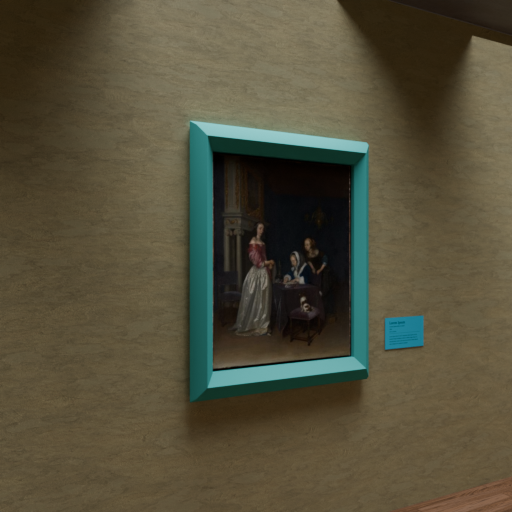} &
        \includegraphics[width=\linewidth]{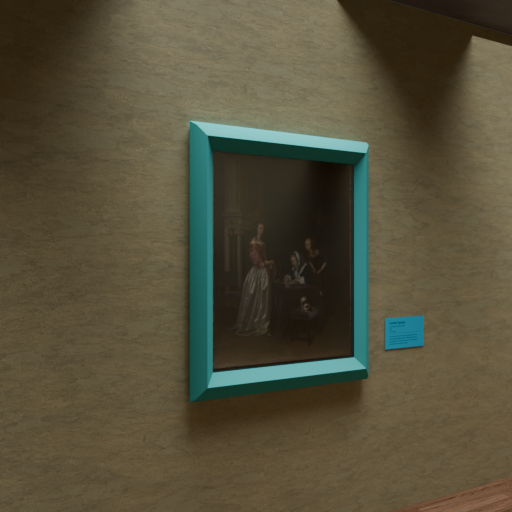} &
        \includegraphics[width=\linewidth]{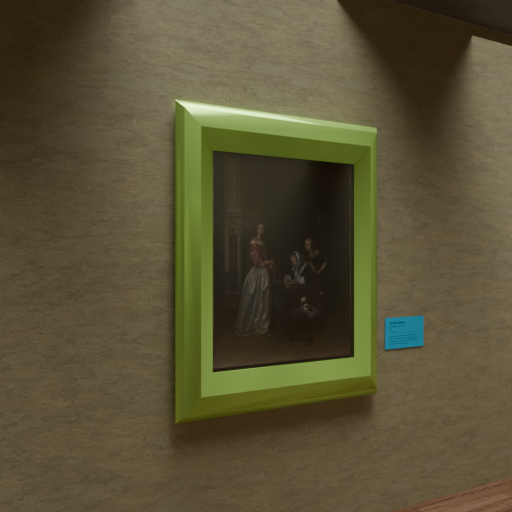} &
        \includegraphics[width=\linewidth]{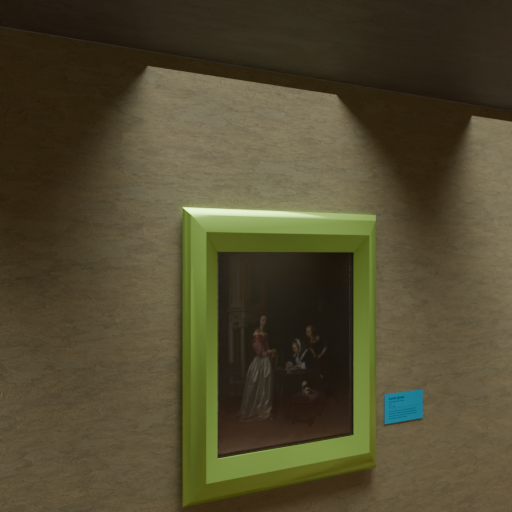} \\
        \includegraphics[width=\linewidth]{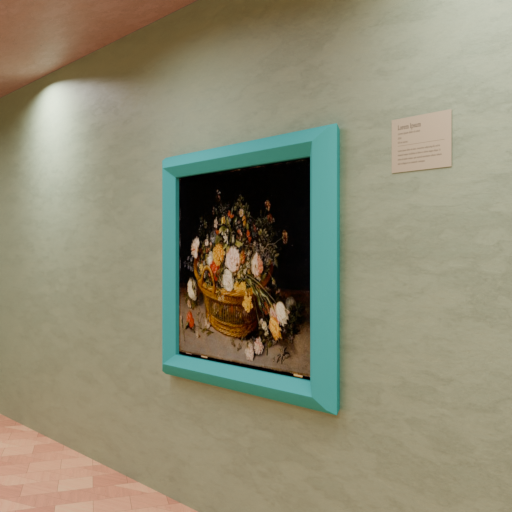} &
        \includegraphics[width=\linewidth]{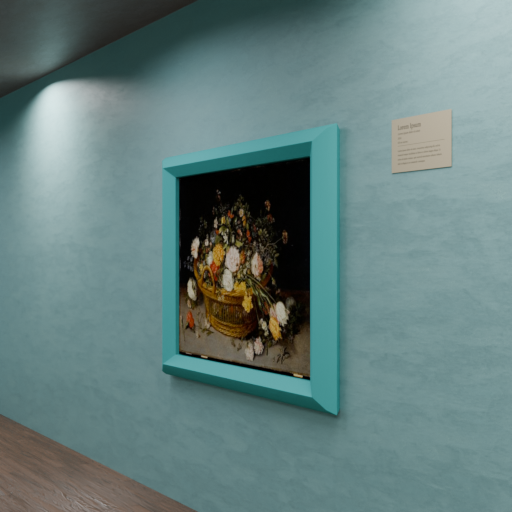} &
        \includegraphics[width=\linewidth]{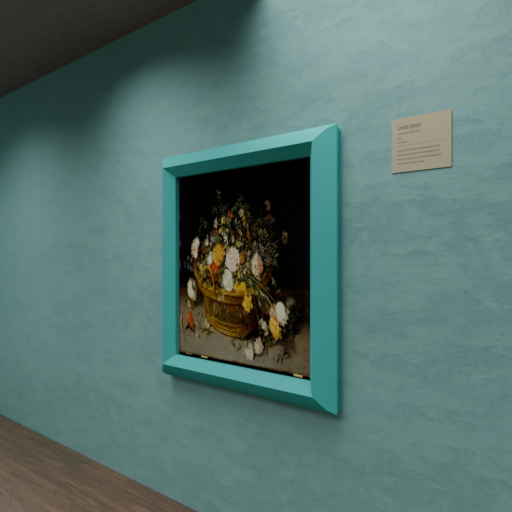} &
        \includegraphics[width=\linewidth]{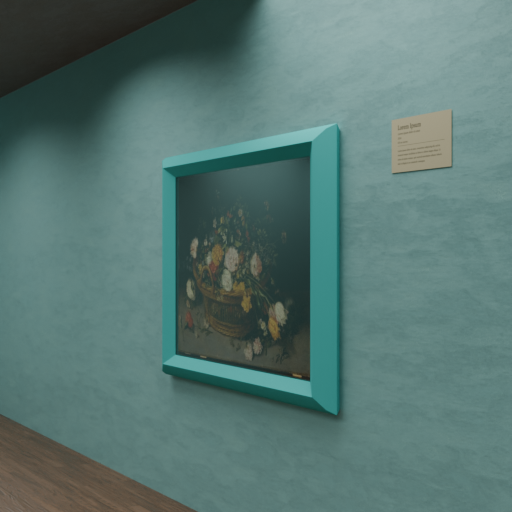} &
        \includegraphics[width=\linewidth]{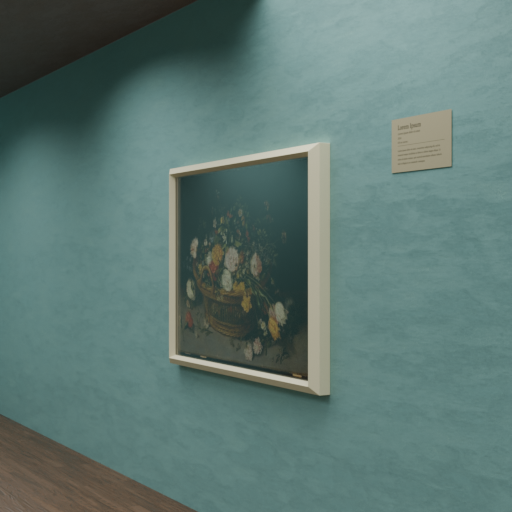} &
        \includegraphics[width=\linewidth]{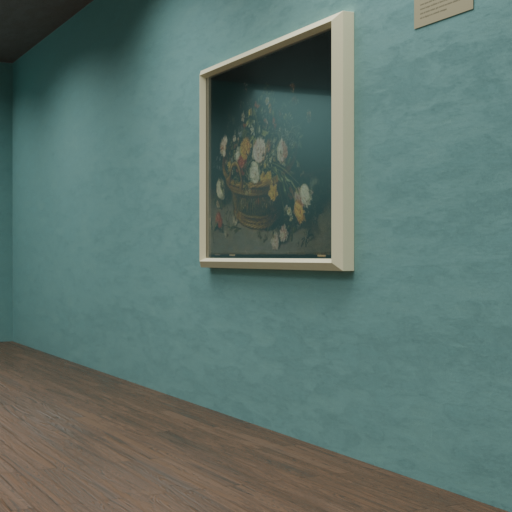} \\
        \includegraphics[width=\linewidth]{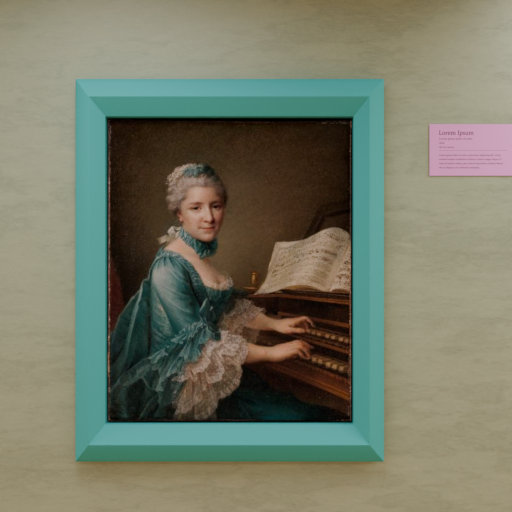} &
        \includegraphics[width=\linewidth]{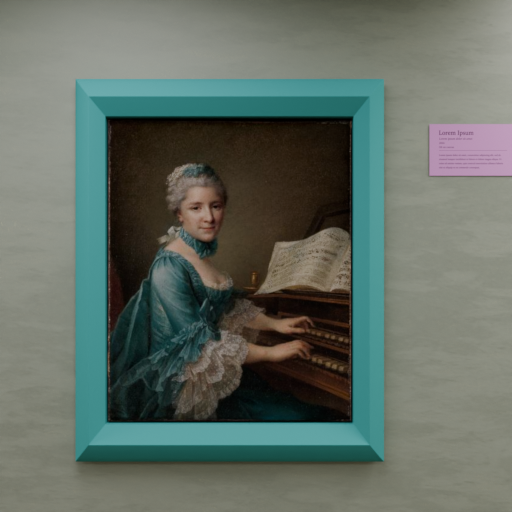} &
        \includegraphics[width=\linewidth]{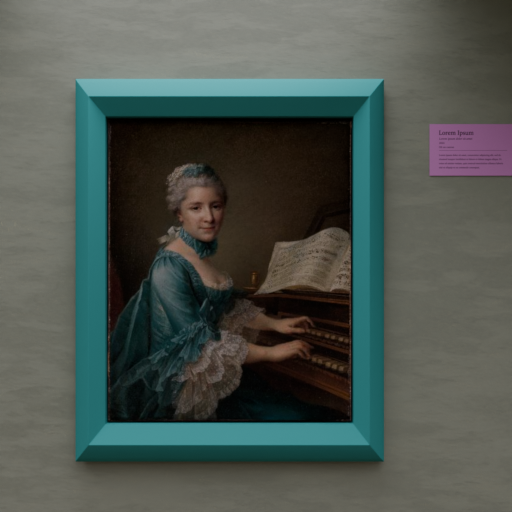} &
        \includegraphics[width=\linewidth]{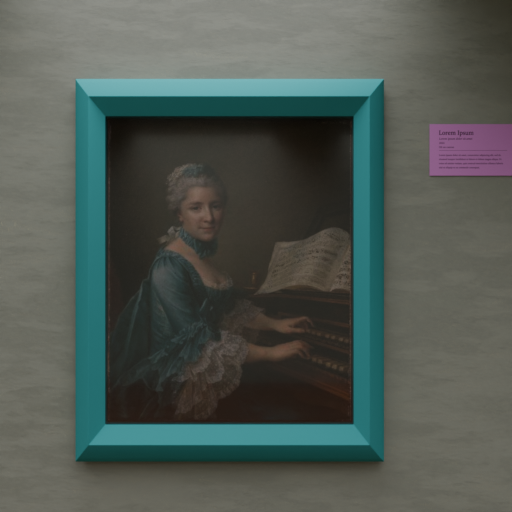} &
        \includegraphics[width=\linewidth]{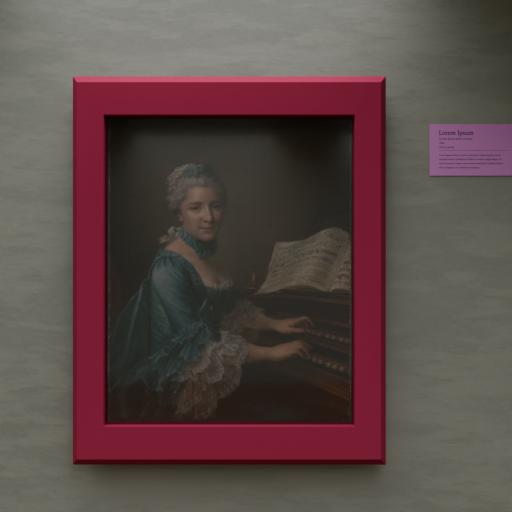} &
        \includegraphics[width=\linewidth]{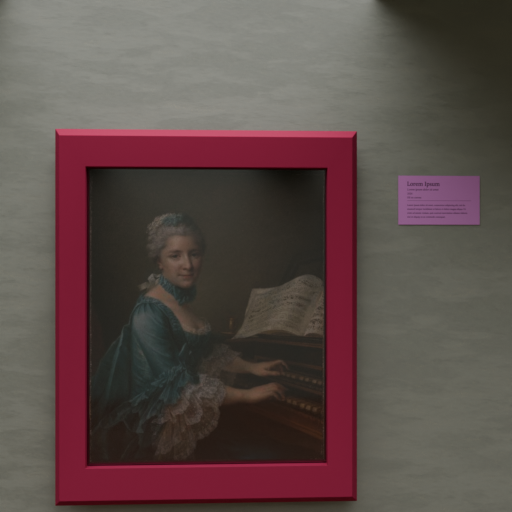} \\
        \includegraphics[width=\linewidth]{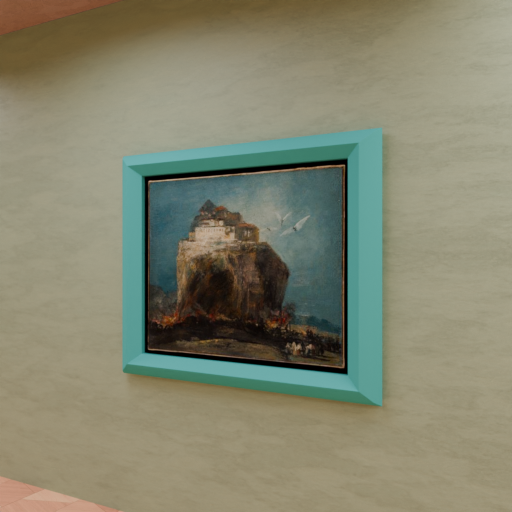} &
        \includegraphics[width=\linewidth]{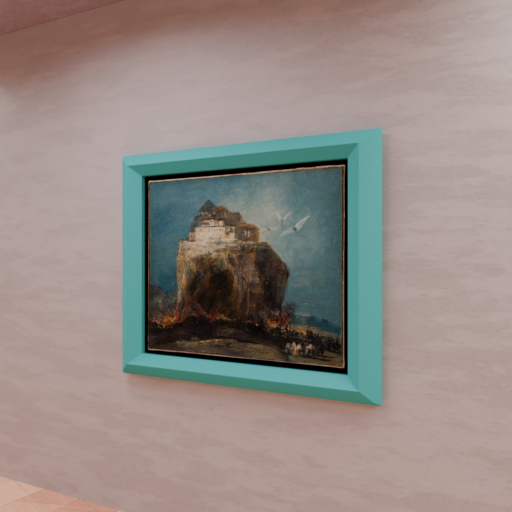} &
        \includegraphics[width=\linewidth]{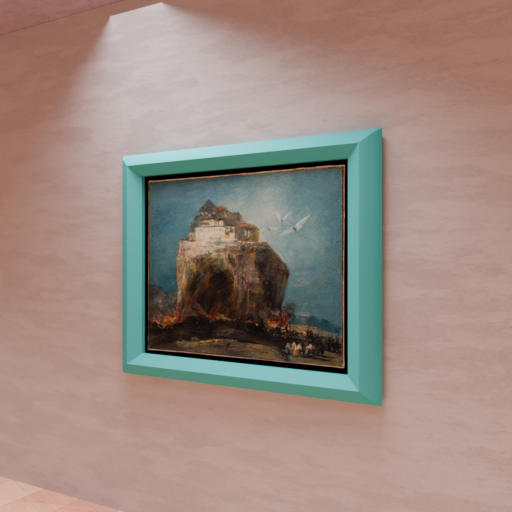} &
        \includegraphics[width=\linewidth]{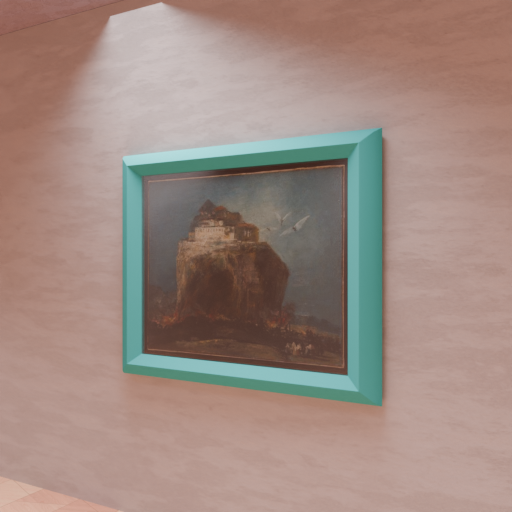} &
        \includegraphics[width=\linewidth]{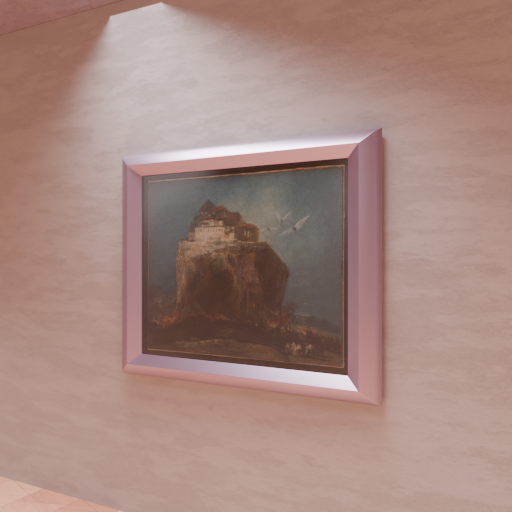} &
        \includegraphics[width=\linewidth]{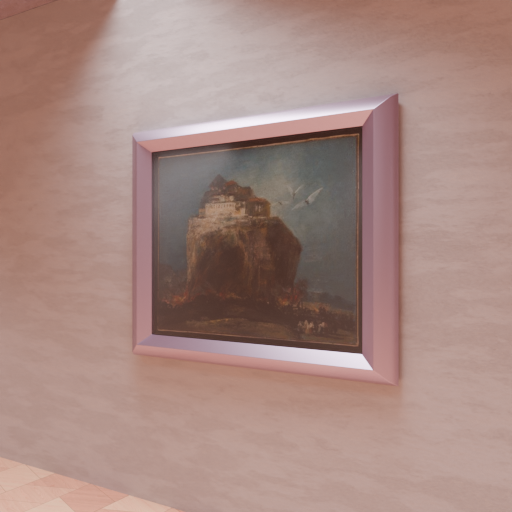} \\
    \end{tabular}
    \caption{Renders from the steps of the dataset ablation. Each row shows the
    same scene under the successive randomization rungs; the rightmost column
    shows the default configuration of \syngallery{}.}
    \label{fig:ablation}
  \end{figure}

  To better understand where the improvement of our synthetically generated dataset lies, we conducted several ablation experiments (see Fig.~\ref{fig:ablation}) with our procedural model to identify which aspects
  contribute to recognition performance. The results in Tab.~\ref{tab:ablation} indicate
  that viewpoint variation has the strongest effect among the tested factors. When the five
  camera views are reduced to a single frontal view, performance drops to $\GAPm$ 68.55,
  close to the all-real baseline. In contrast, scene-level randomizations, including textures, lighting, and frame
  variation, have a more mixed effect and provide only modest gains.
  Rendering at a higher resolution ($1024^2$) also produces only a small improvement. These
  results suggest that, for this task, exposing the model to multiple views of the same
  painting is more important than increasing scene detail or photorealism.

  % --- Table: dataset ablation grid
  \begin{table}[tbp]
  \caption{\textbf{Dataset ablation in the closed painting world.} Models are trained
exclusively on 24{,}490 synthetic renders, varying only the procedural generation
parameters; 148 real painting queries against the 12{,}403-image database, $k/\tau$
chosen via 2-fold cross-validation. Viewpoint coverage is the primary driver of
performance, whereas individual scene randomizations have a more nuanced impact.}
  \label{tab:ablation}
  \centering\footnotesize
  \scalebox{0.9}{
  \begin{tabular}{lcc}
  \toprule
  & \multicolumn{2}{c}{closed painting world} \\
  \cmidrule(lr){2-3}
  training renders & $\GAPm$ & ACC \\
  \midrule
  fixed scene (no randomization) & 73.78 & 75.00 \\
  $+$ textures & 73.94 & 75.00 \\
  $+$ light & 74.42 & 75.68 \\
  $+$ glass ($p{=}0.25$) & 75.32 & \textbf{76.35} \\
  $+$ frame & 72.61 & 73.65 \\
  $+$ camera jitter (default SynGallery) & 74.38 & 75.68 \\
  \midrule
  default, frame variety frozen & 73.91 & 75.00 \\
  default at $1024^2$ & \textbf{75.41} & \textbf{76.35} \\
  default, 3 viewpoints (14{,}694) & 74.31 & 75.68 \\
  default, frontal only (4{,}898) & 68.55 & 70.27 \\
  \midrule
  all-real reference (12{,}403 photos) & 67.18 & 70.27 \\
  \bottomrule
  \end{tabular}
  }
  \end{table}

  % --- Figure: renders of the ablation ladder

\subsection{The Failure of Pixel-Level Augmentation}
A natural alternative is to make the renders more similar to mobile photographs by adding
low-level capture artifacts. We test this by injecting JPEG compression, sensor noise,
resolution loss, and motion blur during training.
Each artifact family is applied stochastically with probability $p=0.5$ per image, so the model always sees a mixture of clean and degraded views rather than a uniformly corrupted training set.

The tested degradations consistently reduce retrieval performance. JPEG compression produces a relatively small decrease of $0.2$ $\GAPm$, while sensor noise combined with resolution loss reduces performance by $1.7$ $\GAPm$, and motion blur causes a larger drop of $2.9$ $\GAPm$. Applying all three degradation families together yields the largest decrease, reducing $\GAPm$ by $3.8$. This outcome is notable because the real visitor queries do contain such artifacts, so matching the training data to the test conditions might be expected to help. A likely explanation is that instance-level painting recognition depends on fine visual evidence, including brushwork, texture, and small compositional details; randomly destroying this evidence during training weakens the learned embedding more than it helps the model cope with degraded queries. An embedding-space audit shows that these artifacts also fail as a proxy for photographic realism: under DINOv3, every artifact family moves the renders further from the real visitor photographs rather than closer. Simulating low-level capture statistics therefore neither imitates the visitor domain nor helps retrieval; the useful training signal in \syngallery{} comes from scene-level variation, especially geometry and illumination.

\section{Conclusion}
\label{sec:conclusion}

We have presented SynGallery, a synthetic dataset and generation pipeline for instance-level artwork recognition under gallery conditions. The central idea is to preserve the identity of each cataloged painting while changing the scene in which it is observed: camera pose, wall placement, frame, lighting, glass, and gallery context. This provides training views that are absent from studio catalog photographs, but characteristic of real visitor queries.

Our experiments show that this synthetic scene-level view variation is a useful training signal for large-scale artwork retrieval. At a matched image budget, training on SynGallery alone outperforms training on the corresponding real studio photographs, and added to the full Met training set it improves the published benchmark protocol without changing the retrieval method or adding real visitor photographs. The ablations further indicate that the gain is driven by view variation, with viewpoint coverage as the strongest single factor, rather than by photorealistic imitation: the renders remain a distinct visual domain in the embedding space, while simulated capture artifacts such as blur, noise, and compression reduce performance.

The current dataset is limited to RGB renderings of paintings and does not yet exploit the full information available from the procedural scene, such as depth, masks, or camera parameters. In future work, we plan to extend the generation to additional object types, richer galleries, stronger occlusions and reflections, and multimodal annotations; synthetic views of distractor classes (statues, furniture) would likely close the remaining gap in open-set retrieval.

\section*{Acknowledgements}
This work was supported in part by the European Union through the EOSC-ARENA project
(AI Research Enhancement through Networked Agents), Grant Agreement No.~101292597 and by PIAST-AI Factory project---PIAST AIF, co-financed by the EuroHPC Joint Undertaking through the European Union's Digital Europe Programme, and by Poland as the Participating Country, under Grant Agreement No.~101250730.

% ---------------------------------------------------------------
% Bibliography
% \clearpage
\FloatBarrier
\bibliographystyle{splncs04}
\bibliography{main}

% ---------------------------------------------------------------
% Appendix (compiled in the arXiv version; omitted from the camera-ready).
\clearpage
\FloatBarrier
\appendix
% llncs restarts the section counter here; give appendix anchors their own
% names so hyperref does not reuse "section.1" (breaks the Appendix A link).
\renewcommand{\theHsection}{appendix.\Alph{section}}
\section{Honest-Tuning Audit}
\label{sec:appendix_honest_tuning}

In our closed-world painting recognition and dataset ablation experiments, we determine the retrieval hyper-parameters $k$ and $\tau$ using 2-fold cross-validation directly on the 148 test queries. Because the test set is relatively small, a potential concern is that this cross-validation strategy might overfit the hyper-parameters to the query set, artificially inflating the reported $\GAPm$ performance.

To verify the integrity of our evaluation, we conduct an ``honest-tuning audit.'' In Tab.~\ref{tab:honest-tuning-audit}, we compare our reported 2-fold cross-validated scores against a leaky ``oracle.'' The oracle score represents the theoretical upper bound achievable by hyper-parameter tuning for this specific dataset. It is obtained by running a grid search over the entire 148-query test set and reporting the absolute maximum score.

The difference between our cross-validated scores and the oracle ceiling is negligible across all data mixes, never exceeding a difference of $0.33$ $\GAPm$. In several instances, the 2-fold score marginally underestimates the oracle, which is expected from a strict data split. This confirms that our tuning strategy does not introduce data leakage, and the performance gains demonstrated by the synthetic renders are driven by the data itself, not by $k/\tau$ optimization.

\begin{table}[h!]
  \centering
  \caption{Honest-tuning audit (closed world). Reported 2-fold cross-validated score vs.\ the leaky ``oracle'' (tune = report on all 148 queries). Every gap is $\leq 0.33$, so the $k/\tau$ choice is not what produces the gains.}
  \label{tab:honest-tuning-audit}
  \scalebox{0.9}{
  \begin{tabular}{lrrr}
    \toprule
    mix & reported (2-fold) & oracle & diff \\
    \midrule
    80:20 & 71.10 & 71.37 & $+0.27$ \\
    60:40 & 71.97 & 71.95 & $-0.02$ \\
    40:60 & 72.27 & 72.21 & $-0.06$ \\
    20:80 & 72.77 & 72.91 & $+0.14$ \\
    0:100 & 73.47 & 73.68 & $+0.21$ \\
    \midrule
    synth125 & 73.61 & 73.70 & $+0.09$ \\
    synth150 & 73.48 & 73.81 & $+0.33$ \\
    synthall & 74.38 & 74.51 & $+0.13$ \\
    \bottomrule
  \end{tabular}
  }
\end{table}

\end{document}